\documentclass[lettersize,journal]{IEEEtran}
\usepackage{amsmath,amsfonts}

 \usepackage[ruled,vlined]{algorithm2e}
\usepackage{algorithmic}

\usepackage{array}
\usepackage[caption=false,font=normalsize,labelfont=sf,textfont=sf]{subfig}
\usepackage{textcomp}
\usepackage{stfloats}
\usepackage{url}
\usepackage{verbatim}
\usepackage{graphicx}
\usepackage{cite}

\usepackage{hyperref}
\usepackage{wasysym}
\newcolumntype{C}[1]{>{\centering\arraybackslash}p{#1}}
\usepackage{fixltx2e}
\usepackage{url}
\usepackage{hyperref}
\usepackage{enumitem}  
\usepackage{adjustbox}
\usepackage{booktabs}  

\usepackage{multirow}
\usepackage[normalem]{ulem}
\useunder{\uline}{\ul}{}

\usepackage{color}
\usepackage{multirow} 
\usepackage{multicol}
\usepackage{makecell}
\usepackage{amsmath,amssymb,amsfonts}
\usepackage{arydshln}
\usepackage{booktabs}
\usepackage{ragged2e}

\begin{document}


\title{FNBench: Benchmarking Robust Federated Learning against Noisy Labels}
\author{Xuefeng~Jiang,
        Jia~Li,
        Nannan~Wu,
        Zhiyuan~Wu,~\IEEEmembership{~Member,~IEEE},
        Xujing~Li,~\IEEEmembership{Student~Member,~IEEE},
        Sheng~Sun,
        Gang~Xu,
        Yuwei~Wang,~\IEEEmembership{~Member,~IEEE},
        Qi~Li,~\IEEEmembership{Senior~Member,~IEEE},
        and Min~Liu,~\IEEEmembership{Senior~Member,~IEEE}
\IEEEcompsocitemizethanks{
Xuefeng Jiang, Zhiyuan Wu and Xujing Li are with the Institute of Computing Technology, Chinese Academy of Sciences, Beijing, China, and also with
the University of Chinese Academy of Sciences, Beijing, China
(E-mail:  \{jiangxuefeng21b, wuzhiyuan22s, lixujing19b\}@ict.ac.cn).
Jia Li is with the Institute of
Information Engineering, Chinese Academy of Sciences, Beijing, China and also with
the University of Chinese Academy of Sciences, Beijing, China. 
(E-mail: lijia1999@iie.ac.cn).
Nannan Wu is with Huazhong University of Science and Technology, Wuhan, Hubei province, China.
(E-mail: wnn2000@hust.edu.cn).
Sheng Sun, Yuwei Wang and Gang Xu are with the Institute of
Computing Technology, Chinese Academy of Sciences, Beijing, China. 
(E-mail: \{sunsheng,ywwang,xugang\}@ict.ac.cn).
Qi Li is with the Institution for Network Sciences and Cyberspace, Tsinghua University, Beijing, China. 
(E-mail:  qli01@tsinghua.edu.cn). 
Min Liu is with the Institute of Computing Technology, Chinese Academy
of Sciences, Beijing, China, and also with the Zhongguancun Laboratory,
Beijing, China. 
(E-mail:  liumin@ict.ac.cn). 
\protect \\
Corresponding Author: Min Liu.

}
\thanks{Manuscript received August 15, 2024; revised May 12, 2025.}

}

\markboth{IEEE TRANSACTIONS ON DEPENDABLE AND SECURE COMPUTING}%
{Shell \MakeLowercase{\textit{et al.}}: A Sample Article Using IEEEtran.cls for IEEE Journals}


\maketitle

\begin{abstract}
Robustness to label noise within data is a significant challenge in federated learning (FL).
From the data-centric perspective, the data quality of distributed  datasets can not be guaranteed since annotations of different clients contain complicated label noise of varying degrees, which causes the performance degradation.
There have been some early attempts to tackle noisy labels in FL.
However, there exists a lack of benchmark studies on comprehensively evaluating their practical performance under unified settings.
To this end, we propose the first benchmark study \textit{FNBench} to provide an experimental investigation which considers three diverse label noise patterns covering synthetic label noise, imperfect human-annotation errors and systematic errors.
Our evaluation incorporates eighteen state-of-the-art methods over five image recognition datasets and one text classification dataset.
Meanwhile, we provide observations to understand why noisy labels impair FL,
and additionally exploit a representation-aware regularization method to enhance the robustness of existing methods against noisy labels based on our observations.
Finally, we discuss the limitations of this work and propose three-fold future directions. 
To facilitate related communities, our source code is open-sourced at \href{https://github.com/Sprinter1999/FNBench}{https://github.com/Sprinter1999/FNBench}.
\end{abstract}

\begin{IEEEkeywords}
Noisy label learning, data quality, federated learning
\end{IEEEkeywords}

\section{Introduction}
\IEEEPARstart{T}{he} pervasiveness of mobile and Internet-of-Things (IoT) devices contributes more than half of the world's internet traffic \cite{fedelc}, which empowers numerous intelligent applications.
In the past years, local data generated by these devices are often directly transmitted from network edge devices to the server and utilized to train a task-specific model at the server following a centralized training paradigm. 
Meanwhile, these mobile and IoT devices (\textit{clients}) are gradually equipped with higher computation capacity.
To protect privacy and exploit the evolving computation capacity to boost model training efficiency \cite{fedSummary}, federated learning  (FL) has emerged as a promising distributed learning paradigm \cite{fedavg}, which allows a subset of clients to train local models individually, and a central server to produce a global model by aggregating client-side local models.
Existing works towards FL have achieved wide success in dealing with four main challenges when deploying a practical FL system including data heterogeneity, system heterogeneity, communication efficiency, and privacy concerns \cite{FedLSR,fedSummary}. 
Meanwhile, it has also been employed for real-world privacy-sensitive scenarios like medical analysis \cite{fedeye,FedNoRo}, IoT malware detection \cite{malwareD} and DDoS Detection \cite{ddos} over distributed networks.

From the data-centric perspective, numerous methods have been proposed from different aspects \cite{fedtrip,fedict,wenke,xiuwen,biad,edgeComputing} to address the data heterogeneity challenge, i.e. non-identically distributed (Non-IID) data across clients, since Non-IID data lead to slow convergence rate and evident performance degradation \cite{fedavg,decorr,xiuwen}.
However, another data-related issue has long been neglected, that is, most existing works are based on the strong assumption that local datasets owned by clients are well annotated \cite{FedLSR}.
In practice, client-side local datasets can easily contain incorrect labels (a.k.a. noisy labels) and it is hard to directly evaluate the data quality of local datasets.
Training on data with noisy labels significantly slows down the model convergence and degrades the final performance of the trained model.
For the client side, actively providing a well-labeled dataset can cost a lot as it involves collecting high-quality annotations that require expensive human intelligence and efforts (e.g. medical scenarios \cite{FedNoRo}).
For the server side, it is hard to quantify the annotation quality of each client's local dataset due to privacy related concerns \cite{fedelc,clc,chen2020focus}.

In recent years, there are attempts \cite{FedLSR,FedNoRo,RFL,fedrn,rfa,median,fedelc,co-teaching} that can be considered to tackle distributed noisy clients.
These works often conduct experiments under very different experimental settings like different base model selection, different datasets, varying fundamental hyper-parameter selection and data partitioning strategies.
For example, some early works like \cite{FedLSR,rhfl,RFL} mainly consider the unpractical independent and identically distributed (IID) client data.
Meanwhile, previous works \cite{rhfl,dualoptim,FedNoRo} often consider relatively limited label noise patterns and mainly focus on the image recognition task, which are hardly representative and thorough.
For federated noisy label learning (FNLL) methods, previous works often compare with different methods and lack comparisons with other concurrent FNLL methods.
Therefore, there lacks a benchmark study on meticulously reviewing existing methods and systematically understand their actual performance against different data partitioning and complicated label noise patterns. 
In addition, rare works aim to further explore underlying potential reasons why noisy labels bring degradations to the convergence and final performance of trained models.

In this work, we retrospect the latest advance from recent literature and implement an easy-to-use benchmark framework \textit{FNBench} to conduct a comprehensive benchmark study of related methods under unified experimental settings. 
\textit{FNBench} incorporates eighteen related state-of-the-art methods and three distinct label noise patterns including synthetic label noise, imperfect human-annotation errors and systematic errors.
We provide an early empirical investigation into the robustness of existing methods for the natural language processing task.
The experimental results and our conclusions provide insights for the further development of future feasible methods. 
Meanwhile, through our explorations to understand the performance degradation caused by noisy labels, we observe the distribution of learned representations also evidently degrade, which is often referred by dimensional collapse \cite{directCLR} in representation learning.
To this end, we propose to exploit a representation-aware regularization technique to mitigate this dimensional collapse.
By combining this representation-aware technique with existing methods, we can generally further improve the robustness of these methods against the complicated label noise issue.
Last, to benefit related communities, our code base is publicly available. 
Researchers can easily utilize our code base to try different experimental settings and diverse label noise patterns for the evaluation of existing methods or implement their new proposed methods.

Specifically, our main contributions can be summarized as follows:
\begin{itemize}[leftmargin=0.3cm]
\item We carefully review related literature about approaches to tacking noisy labels in the context of FL. Meanwhile, we provide observations to inspect why noisy labels impair FL from two underlying perspectives called deep network memorization effect and dimensional collapse.
\item We provide an open-source benchmark evaluation framework \textit{FNBench} which supports eighteen state-of-the-art methods with diverse experimental settings to better facilitate future studies.
\item We conduct extensive experiments to evaluate these methods over three distinct label noise patterns. Through extensive empirical evaluations, we draw diverse conclusions to guide future studies.
\item Based on our observations and analysis on dimensional collapse, we further propose to exploit an effective representation-aware regularization method to enhance the performance of existing methods. Additionally, we discuss the limitations of this study and carefully provide three-fold directions for future studies to explore.
\end{itemize}

The remainder of this work has been organized in the following way. Section \ref{RW} discusses related literature.  Section \ref{prelin} and \ref{benchmarkDesign} present the general FL pipeline and our benchmark design, then Section \ref{observations} analyzes the underlying two phenomena named memorization effect and dimensional collapse to provide observations to understand why noisy labels pose challenges in FL. Section \ref{sec:exp} reports empirical evaluations under extensive experimental settings and draw corresponding conclusions. We discuss limitations and future directions and then summarize this work in Section \ref{sec:discuss} and \ref{sec:conclusion}.

\section{Related Works}
\label{RW}
In this section, we first outline FL and the pervasive existence of noisy labels.
Then we review noisy label learning in both centralized learning and FL.
Finally we introduce some previous benchmark studies towards FL to better present the motivation of this work.

\subsection{General Federated Learning (FL)}
FL provides an opportunity to leverage distributed local datasets to derive a generalizable model. 
One key challenge in such distributed datasets is that data distributions in different clients are usually non-identically distributed (non-IID), and many works are tackling this issue from various aspects \cite{fedtrip,fedict,qinbinCrossSilo,crac,noniidsurvey,wenke,xiuwen,biad,smartphone,jryInfocom,tanhao,fedlf}. 
FedAvg \cite{fedavg} is the pioneering FL work which aggregates the local trained models based on the size of local datasets. 
FedProx \cite{fedprox} proposes a proximal term on the client side to restrain the local model parameters from deviating too much from the received global model parameters of the current communication round.
FedExP \cite{FedExP} speeds up FedAvg via extrapolation on the server side.
However, most literature on FL has an underlying assumption that each client is willing to provide a clean local dataset with well-annotated labels, which is not practical in real-world scenarios. 

\subsection{The Source of Noisy Labels}
Generally, high-quality labeled data require expensive human intelligence and cost, thus, local datasets tend to be not well-annotated without supervision. Herein we list reasons why local datasets contain noisy labels:

\textbf{Auto-generated labeling.} There are some common low-cost auto label generating methods \cite{FedLSR}. A client can collect images as its local dataset over online websites and filter keywords in the context as labels \cite{mopro,clothing1m}. It can also exploit machine-generated labels using an imperfect pretrained model on a small-scale pre-collected dataset to predict possible labels for its local dataset \cite{DBLP:journals/ijcv/KuznetsovaRAUKP20}.

\textbf{Limited expert knowledge.} Collecting annotations can require high-level human expertise and efforts such as medical scenarios \cite{FedNoRo}.
Even widely-used datasets like CIFAR-10 \cite{cifar} and Amazon-Reviews \cite{amazon} contain noisy labels \cite{FedLSR,pervasiveerrors}.
Given limited budgets to annotate the dataset, clients can resort to distribute the labeling task to not highly-skilled annotators via crowd-sourcing to save expense \cite{robustfed}.

\textbf{Free rider and malicious attacker.} If clients are not given enough rewards to participate in FL \cite{incentives}, their motivation to provide well-annotated local datasets diminishes. 
Meanwhile, it is hard to access user data and evaluate the quality of local data due to privacy \cite{gdpr}.
Clients tend to provide local datasets of coarse and low-quality annotated datasets (i.e. free-rider) and even worse, some clients may intentionally become attackers and provide bad and unsafe local datasets \cite{robustfed,iotj}. 

Training on imperfectly labeled datasets leads to an inferior and unsafe model.
Therefore, it is of significance to explore methods to train robust models over data with noisy labels.

\subsection{Noisy Label Learning (NLL)}
\label{nllsec}
An early study \cite{memorization} points out DNNs are capable of overfitting data with noisy labels, which causes a direct generalization degradation.
In centralized scenario \cite{NLL}, existing works can be roughly divided into two approaches. 
One main approach is conducting sample selection \cite{fedrn}. Co-teaching \cite{co-teaching} and its derivative Co-teaching+ \cite{co-teaching+} are two pioneering methods which maintain two peer networks with different initialization and each network selects training samples with designed loss-seperation protocols for another peer network, utilizing different discrimination ability of peer networks.
Another approach focuses on robust training techniques or robust loss design. 
JointOptim \cite{jointopt} and SELFIE \cite{selfie} pioneer to alternately update model parameters and labels by averaging the deep network's output predictions generated in previous training epochs.
By this way, part of noisy labels can be gradually optimized to the right labels.
Regarding robust loss design, some works analyze the imperfection of traditional cross entropy (CE) loss in the existence of noisy labels and propose their robust loss design \cite{symmetricCE,MAE,GCE}.
Wang et al. \cite{symmetricCE} propose model predictions can reflect the ground truth label to a certain degree, and incorporate model prediction into supervision to derive a robust loss function Symmetric CE.
DivideMix \cite{Dividemix} integrates multiple techniques including Co-teaching, MixUp \cite{mixup} and MixMatch \cite{mixmatch}. 

We implement several NLL methods which can be directly and easily combined with the classic FL pipeline since other methods like \cite{shengL} require some unrealistic assumptions in FL as analyzed in \cite{FedLSR}. For more information about noisy label learning, please refer to \cite{NLL,NLLnara}.

\subsection{Federated Noisy Label Learning (FNLL)}

FNLL aims to directly tackle noisy labels in FL, which is less studied in the overall community of FL.
As far as we investigate, RFL \cite{RFL} is the first work to directly deal with noisy labels and collects local classwise feature centroids to form global classwise feature centroids as extra global supervision.
The intuition is that the local classwise centroids should be close to global classwise centroids.  
FedLSR \cite{FedLSR} points out RFL's global supervision carries sensitive client privacy, and proposes a client-side three-fold local self-regularization method  to mitigate over-fitting noisy local datasets.  
FedRN \cite{fedrn} considers the complex various Non-IID and various extents of noisy data within different client-side datasets and proposes to maintain a client model pool at the server side. 
Each participating client can be associated with several more reliable neighbors maintained in that pool, which are other client models that have similar data class distributions or minimal label noise.
FedCorr \cite{fedcorr} introduces multiple cascading training stages while it is hard to divide total rounds for each stage.
Meanwhile, FedCorr's multi-step LID score calculations require high computational complexity \cite{icme}. 
FedNoRo \cite{FedNoRo} provides a two-stage approach and explicitly divides all clients into the clean group and the noisy group via fine-grained statistics, requiring the loss distributions of all classes to depict each client and two-component Gaussian mixture model \cite{gmm}. 
FedNoRo designs different training objectives for clean clients and noisy clients. 
Based on FedNoRo, FedELC \cite{fedelc} pioneers to correct the underlying labels via back propagation. 
There are some FNLL works like RHFL \cite{rhfl} and FedNed \cite{fedned} that largely rely on a pre-collected perfectly labeled proxy dataset on the server side, which is a hard assumption \cite{FedLSR,icme}. 
FedELR \cite{fedelr} forces local models to stick to their early training phase via an early learning regularization, however, it requires each client's full participation in FL.
Since these methods are not compatible and not flexible with conventional FL settings, we did not include them into evaluation considering fair comparisons.

\subsection{Byzantine Robust Federated Learning (Robust FL)}
\begin{figure}[htbp]
\centering
\includegraphics[width=0.9\linewidth]{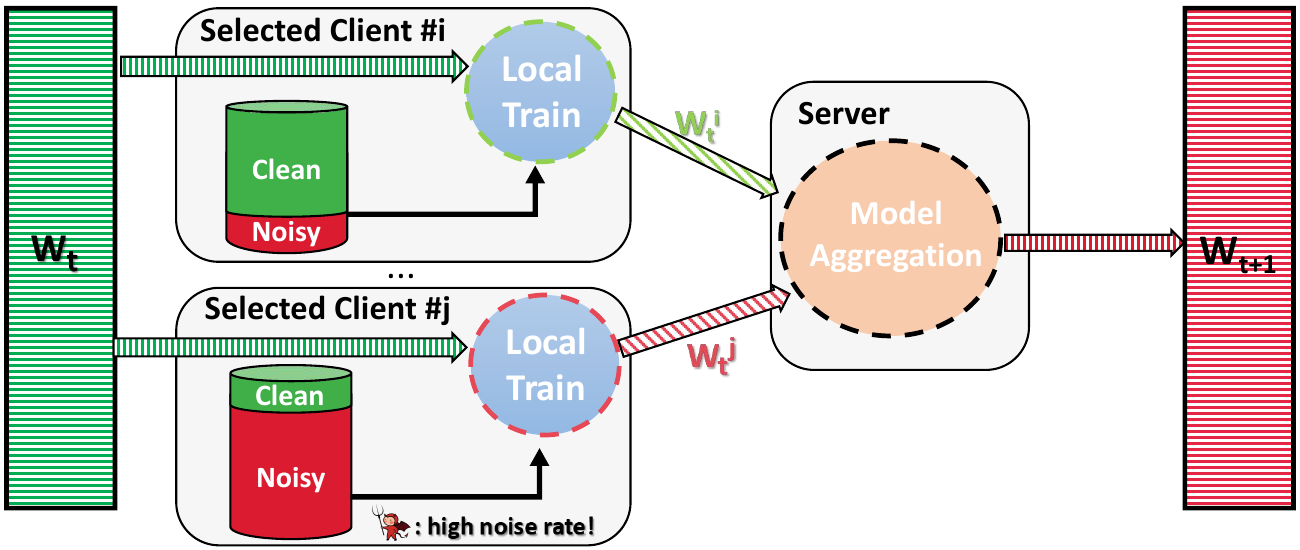}
\caption{Clients in FL are of various noise rates. Clients of higher noise rates tend to yield worse models and further affect the convergence and final performance of the aggregated global model. Many Robust FL methods aim to mitigate the negative effects caused by these high-noise clients via more cautious model aggregation at the server side.}
\label{fig:rob}
\end{figure}

Clients of high label noise rates can be seen as byzantine clients in FL \cite{Agr} which bring data poisoning attacks \cite{robustfed}, as illustrated in Figure \ref{fig:rob}. 
To tackle these unreliable clients, byzantine robust federated learning (Robust FL) methods have been proposed.
Most existing methods aim to increase the robustness of FL by conducting more robust aggregation \cite{rfa}.
Median \cite{median}  aggregates the median values of the clients' models instead of the weighted averaged value in FedAvg \cite{fedavg}. In this way, extreme bad weights can affect less to the global model.
TrimmedMean \cite{TrimmedMean} removes the largest and smallest weights for each parameter of selected clients' models, and computes the mean of the remaining parameters.
Krum \cite{krum} firstly computes the nearest neighbors to each local model. Then, it calculates the sum of the distance between each client and their closest local models. 
Finally, it selects the local model with the smallest sum of distance as the global model. 
One more recent method RFA \cite{rfa} proposes the geometric median with Weiszfeld-type optimization algorithm.

\subsection{Motivation}
\label{MOTIVATION}
There are existing representative benchmark or experimental studies in FL, but there still lacks a comprehensive investigation on the label noise issue in FL.
Li et. al. \cite{qinbinCrossSilo} studies four FL methods' performance in various Non-IID scenarios. 
Some works aim to provide an open-sourced systematic testbed for evaluations \cite{jiachen,fedml,leaf,baochun,fedlab}.
For the FNLL research, we notice a concurrent work \cite{Fednoisy} solely studies FedAvg's performance under various noisy scenarios, but lacks experimental comparisons over other methods.
Meanwhile, existing works compare methods under un-unified settings, which motivates us to provide an experimental study under an unified and fair evaluation framework.
In addition, previous FNLL research works mainly considers the vision recognition task and few considers the natural language related task in the presence of noisy labels.
Considering limitations of previous works, to better facilitate future studies, our main motivation is to derive a systematization of related knowledge and provide the first comprehensive experimental study of state-of-the-art methods under an unified open-sourced evaluation framework.

\section{Preliminary}
\label{prelin}
We firstly present related notations to showcase the general FL pipeline, then conduct preliminary experiments on CIFAR-10 (see Section \ref{ExpSet}) to further analyze the underlying two phenomena (i.e. memorization effect and dimensional collapse) to provide insights into understanding why noisy labels impair FL.

Let $\mathcal{D}=\left\{\left(x_{i}, y_{i}^{*}\right)\right\}_{i=1}^{|\mathcal{D}|}$ with $M$ classes denote the full training dataset, where $x_i$ denotes a training sample and $y_i^{*}$ denotes the given label which can be corrupted to an incorrect label (i.e. noisy label). 
In FL, suppose there are $N$ clients, denoted as $c_1$, ..., $c_N$. The full dataset $\mathcal{D}$ is distributed to all clients.
Each client $c_k$ maintains a local imperfect dataset with $n_k$ samples denoted as $\mathcal{D}_{k}=\left\{\left(x_{i}^{k}, y_{i}^{k}\right)\right\}_{i=1}^{n_{k}}$.
We use $w_t$ and $w_t^k$ to denote the global model and the local trained model of the client $c_k$ in the current communication round $t$. 
As shown in Figure \ref{fig:case}, we use the classic FL \cite{fedavg} to showcase the general FL pipeline. Each communication round $t$ generally contains 4 steps.
\begin{itemize}[leftmargin=0.3cm]
    \item \textbf{Step 1}: The server firstly sends the global model $w_t$ to the selected client set $S_t$ to participate in FL.
    \item \textbf{Step 2}: Each selected client $c_k$ updates the local model $w_t^k$ with its local dataset $D_k$.
    \item \textbf{Step 3}: Afterwards, all local updated models are sent back to the server.
    \item \textbf{Step 4}: The server aggregates the received local models of $S_t$ as the global model $w_{t+1}$ in next round $t+1$ via $w_{t+1}=\sum_{k \in S_t} \frac{n_{k}}{n_{S_t}} w_{t}^k$, where $n_{S_t}$ denotes the total sample number of all selected clients in the current round $t$. 
\end{itemize}
Unlike traditional distributed training \cite{limu}, clients locally update their model $w_t^k$ with multiple local epochs $e$, which decreases the total number of communication rounds between clients and server and is more communication-efficient.
\begin{figure}[htbp]
    \centering
    \includegraphics[width=0.9\linewidth]{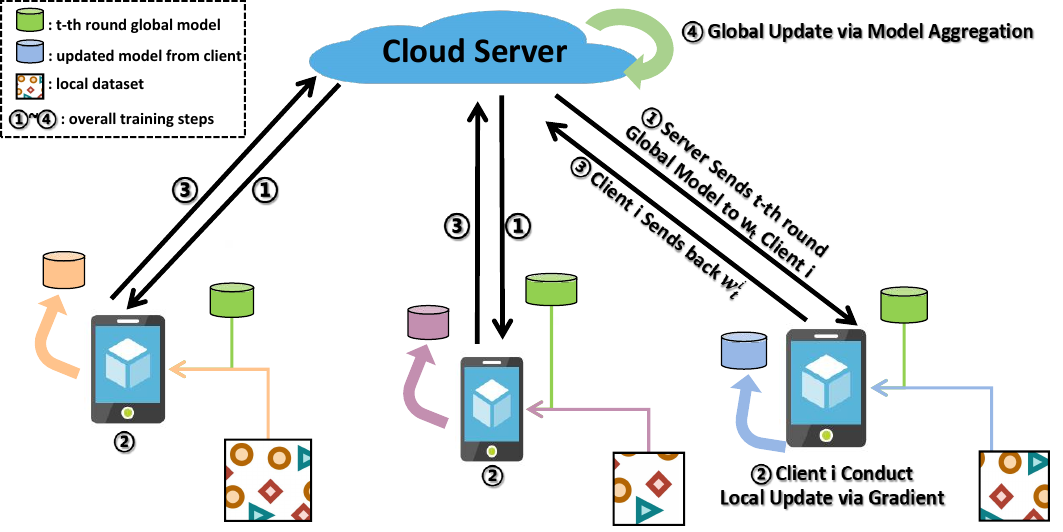}
    \caption{The training paradigm of FL \cite{FedLSR,fedavg}. Local datasets are often Non-IID, and they can contain data with noisy labels which hampers the global model's convergence and final performance.}
    \label{fig:case}
\end{figure}

\section{Benchmark Design}
\label{benchmarkDesign}
In this section, we introduce our benchmark settings and details of evaluated baselines. 

\subsection{Label Noise Patterns}
\label{sec:datasets}
To cover more diverse label noise patterns, we consider three types of label noise in this study:
\begin{itemize}[leftmargin=0.5cm]
    \item \textbf{Synthetic label noise}: Most existing NLL and FNLL works conduct their main experiments on datasets with synthetic noisy labels, which is a mainstream selection of noisy dataset generation in previous works.
For this label noise type, we use two image datasets CIFAR-10/100 \cite{cifar} and one text dataset AGNews \cite{agnews} with synthetic noisy labels. To our best knowledge, main existing works focus on image recognition tasks while rare works investigate the robustness on natural language processing related tasks.
    \item \textbf{Imperfect human annotation errors}: Only experimenting on synthetic noisy labels is far from enough. In this study, we conduct experiments on CIFAR-10/100-N datasets \cite{cifarn} provided by Amazon Mechanical Turk, which reflects the real-world annotation errors caused by human annotators\footnote{There are several versions of CIFAR-10/100-N annotations \cite{cifarn}. We utilize CIFAR-10-N-Worst and CIFAR-100-N-Fine, which have highest noise rates.}. To our best knowledge, we are the first to widely benchmark and evaluate existing methods' robustness on CIFAR-10/100-N datasets reflecting real-world human annotation errors, inspired by \cite{cifarn,chaos}.
    \item \textbf{Systematic errors}: Systematic label noise is often caused by imperfect data and annotation collection process. In this study, we conduct experiments on real-world large-scale dataset Clothing1M \cite{clothing1m}, which collects one million images from famous online shopping websites and filters context as annotated labels. 
\end{itemize}

The statistics of these datasets are shown in Table \ref{tab:dataset}. In the following, we detailedly discuss the generation of noisy labels for these three scenarios. 
\begin{table}[htbp]
  \centering
    \caption{Datasets and corresponding base models.}
  \label{tab:dataset}
  \begin{adjustbox}{width=\columnwidth,center}
\begin{tabular}{ccccccc}
\midrule \midrule
\textbf{Noise Type}                   & \multicolumn{3}{c}{Synthetic}& \multicolumn{2}{c}{Human-annotation Errors} & Systematic Errors \\ \midrule
\textbf{Dataset}                      & CIFAR-10        & CIFAR-100        & AGNews& CIFAR-10-N             & CIFAR-100-N           & Clothing1M       \\ \midrule
\textbf{Modality}      & Image              & Image              & Text & Image                   & Image                & Image               \\
\textbf{Size of $\mathcal{D}$}        & 50,000          & 50,000           & 120,000& 50,000               & 50,000              & 1,000,000        \\
\textbf{Size of $\mathcal{D}_{test}$} & 10,000          & 10,000           & 7,600& 10,000               & 10,000              & 10,000           \\
\textbf{Classes $M$}      & 10              & 100              & 4& 10                   & 100                 & 14               \\
\textbf{Base Model}                   & ResNet-18       & ResNet-34        & FastText& ResNet-18            & ResNet-34           & ResNet-50        \\ \midrule 


\end{tabular}
  \end{adjustbox}
\end{table}

\subsection{Data Partitioning}
For the \textbf{synthetic label noise scenarios} which are the most common experimental scenarios in previous works, to jointly partition Non-IID data with noisy labels, our evaluation considers 3$\times$4 settings (Three Non-IID settings and four label noise settings). 
Detailedly, we merge two related settings following below steps sequentially:
\begin{itemize}[leftmargin=0.5cm]
    \item Non-IID partioning: Referring to \cite{noniidsurvey, fedrn}, we apply two commonly-used ways to distribute whole data to $N$ clients: 
    \textit{(i) Sharding (Pathological)}: Training data are sorted by its class labels and divided into $N*S$ shards, which are randomly assigned to each client with an equal number $S$ of data shards. 
   \textit{ (ii) Dirichlet Distribution}: Each client is assigned with training data drawn from the Dirichlet distribution with a concentration parameter $\beta$. In this study, we assign $\beta$ with $0.5$ and $1.0$.
    \item Synthetic Label noise: Referring to \cite{fedrn, FedLSR, RFL, fedcorr, FedNoRo,NLL,icme}, we inject synthetic label noise via typical protocols, illustrated in Figure \ref{fig:case}:
    \textit{(i) Symmetric Label Noise} \cite{symmN}: A true label is flipped into all possible classes with equal probability.
    We inject varying levels of label noise into local datasets so clients have different noise rates.
    For instance, in table \ref{tab:cifar10} noise of $0.0-0.8$ means that the noise rate increases linearly from $0.0$ to $0.8$ as the client's index increases.
    \textit{(ii) Pairflip Label Noise} \cite{co-teaching}: A true label is flipped into only a certain type of class.  This is also named by asymmetric (Asym.) label noise.
    \textit{(iii) Mixed Label Noise} \cite{fedrn}: Besides above two common noise generation patterns, inspired by \cite{fedrn}, we further conduct evaluation under a more complex scenario named mixed noise where two different types of label noise are injected into all clients. Symmetric label noise is applied to half of clients, while pairflip label noise is applied to another half.
\end{itemize}

\begin{figure}[htbp]
    \centering
    \includegraphics[width=0.9\linewidth]{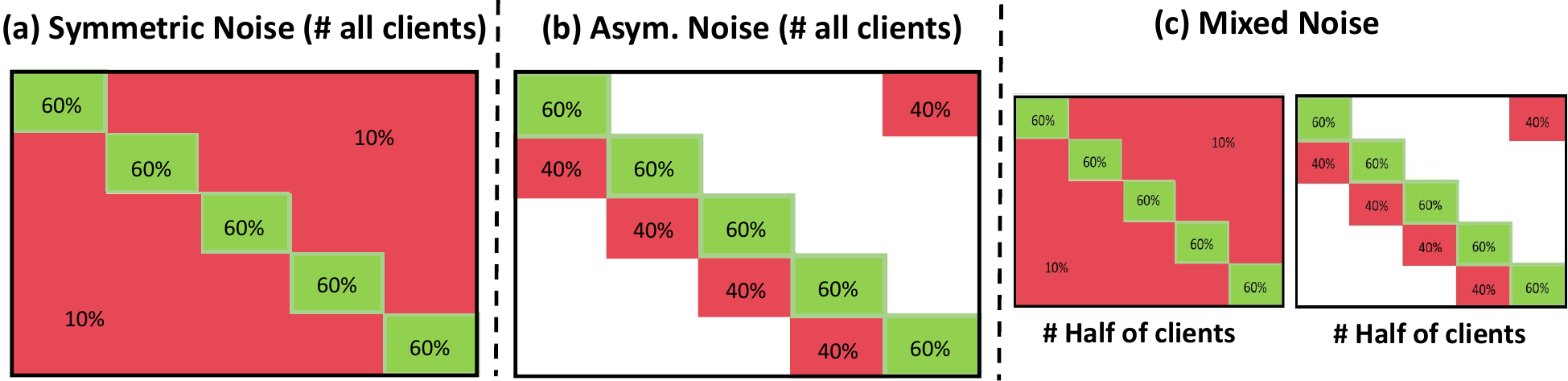}
    \caption{Noise transition matrices of different synthetic noise types (using 5 classes as example). Green grids denote the percentage that is correctly labeled to the ground truth class while the red denotes the wrongly labeled percentage to other classes. }
    \label{fig:case}
\end{figure}

\begin{figure}[htbp]
    \centering
    \includegraphics[width=0.8\linewidth]{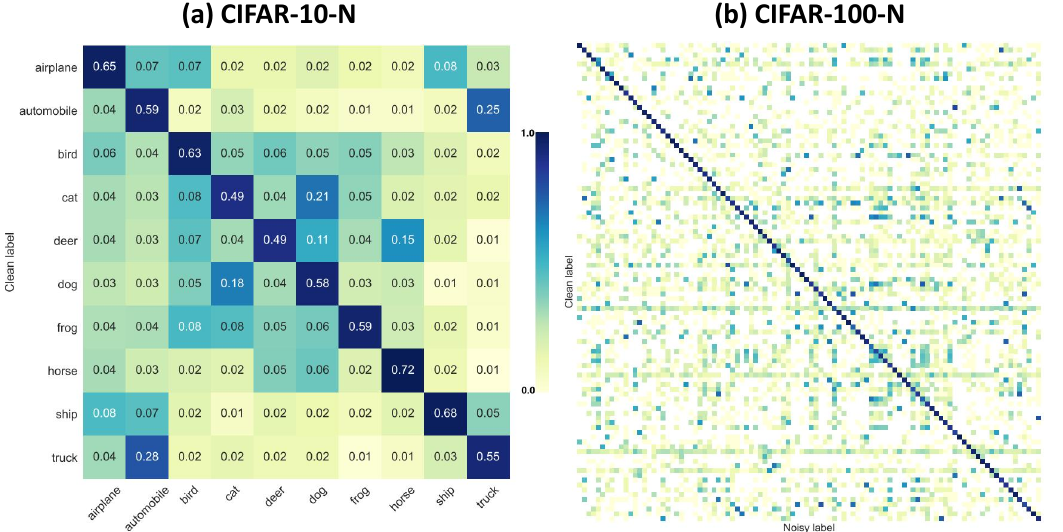}
    \caption{Noise transition matrices for CIFAR-10/100-N. Human-annotation noise transition patterns are more complex than synthetic noise transition patterns in Figure \ref{fig:case}. For more information, please kindly refer to \cite{cifarn}.}
    \label{fig:cifarn}
\end{figure}

For \textbf{human-annotated label noise scenarios}, following previous work \cite{chaos}, we first conduct the data partitioning, and then load human annotations to replace the original labels. The overall noise rate of CIFAR-10-N is 40.208\% while the noise rate for CIFAR-100-N is 40.200\%. The noise transition matrices for these datasets are visualized in Figure \ref{fig:cifarn}.

For \textbf{systematic label noise scenarios}, the annotated labels of Clothing1M \cite{clothing1m} are obtained from several online shopping websites by simply filtering the related web context. The overall noise level is reported to be approximately 39. 46\%.
Following previous works \cite{FedLSR,fedelc}, we randomly divide 1 million images into 100 partitions with the equivalent number of samples, and set them as local clients' datasets. Note that in this partitioning, each client's dataset naturally follows the Non-IID fashion with unstructured label noise, and we use the 10k dataset with correct labels for testing. 

\subsection{Evaluated Baselines}
\label{sec:baselines}
We implement eighteen related methods as baselines from four research groups (discussed in Section \ref{RW}). 
Unless otherwise specified, most hyper-parameters of these methods are configured favorably in line with the original literature. 
We refer to related open-sourced codes to conduct our experiments.
For convenience, we clarify the important hyper-parameter setup of all methods as follows:
\begin{enumerate}[leftmargin=0.5cm]
    \item General FL methods: We compare FedAvg \cite{fedavg},  FedProx \cite{fedprox} and FedExP \cite{FedExP}. The $\mu$ of FedProx is mildly set to 0.1 and $\epsilon$ of FedExP is $1e^{-3}$. 
    \item Robust FL methods: We compare Krum \cite{krum}, Median \cite{median}, Trimmed Mean \cite{TrimmedMean} and RFA \cite{rfa}. Krum and Trimmed Mean require to know the upper bound of the ratio of compromised or bad clients (denoted as $\kappa$), empirically  $\kappa$ is selected from $0.1-0.3$ \cite{Byzantium}. We select $\kappa$ to $0.3$ since  $\kappa$ is actually agnostic since different participants hold various and unknown extent of label noise in their local datasets. 
    \item Noisy label learning (NLL) methods:  We evaluate JointOptim \cite{jointopt}, Symmetric CE \cite{symmetricCE}, SELFIE \cite{selfie}, Co-teaching \cite{co-teaching}, Co-teaching+ \cite{co-teaching+} and DivideMix \cite{Dividemix}. These methods are easy to combine with classic FedAvg's pipeline discussed in Section \ref{prelin}. The original $\alpha$ is set to 0.1 and $\beta$ is set to $1.0$ for Symmetric CE. We report the better F1 score of two peer networks of DivideMix, Co-teaching and Co-teaching+.
    \item Federated noisy label learning (FNLL) methods: We evaluate RFL \cite{RFL}, FedLSR \cite{FedLSR}, FedRN \cite{fedrn}, FedNoRo \cite{FedNoRo} and FedELC \cite{fedelc}. The warm-up rounds for RFL, FedLSR, FedRN, SELFIE \cite{selfie} are mildly set to 20\% of total communication rounds. The $\gamma_e$ and $\gamma$ of FedLSR are set to $0.3$ and $0.4$. The reliable neighbor number is set as 1 for FedRN. The forget rate of Co-teaching, Co-teaching+ and RFL is mildly set to $0.2$ in our study.
\end{enumerate}

\section{Observed Phenomena}
\label{observations}
\subsection{Memorization Effect}
Learning on data with noisy labels is well studied in the centralized learning.
As introduced in \ref{nllsec} and \cite{overfit}, deep neural networks (DNNs) with a large number of parameters can lead to overfitting data with noisy labels.
In detail, \cite{memorization} proposes deep network memorization effect, which indicates that deep networks tend to fit correctly-labeled data before noisy-labeled data during training.
Therefore, there often exists a short warm-up period during which the model generalization performance (reflected by test accuracy) increases at the early learning process but then gradually decreases due to negative effects caused by noisy labels.
This effect also inspires many follow-up methods like Co-teaching and RFL \cite{co-teaching,RFL}. 

\begin{figure}[htbp]
    \centering
    \includegraphics[width=\linewidth]{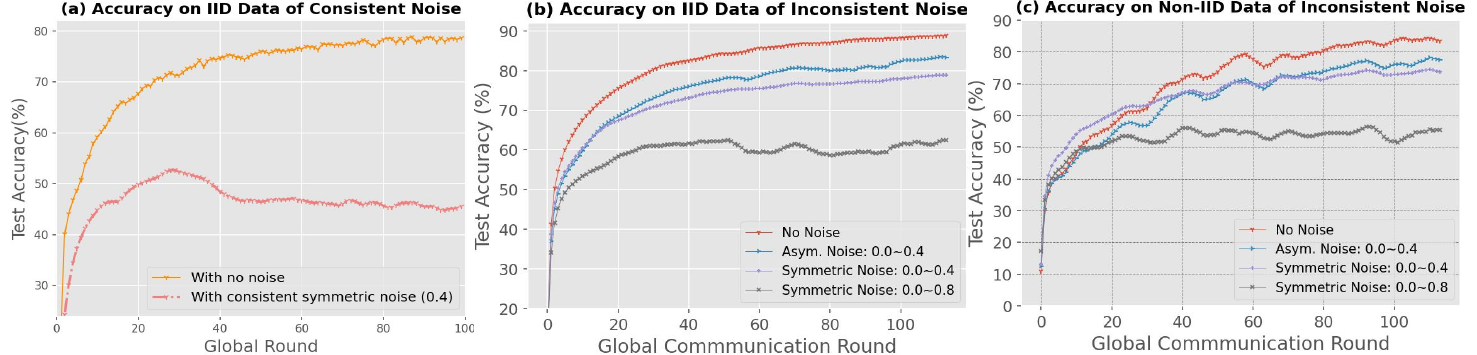}
    \caption{Test accuracy curves on both IID and Non-IID data. (a) illustrates the ideal case where clients have the same noise rate on IID data, which aligns the settings of FedLSR \cite{FedLSR}. Note that the noise rates across clients in (b) and (c) are inconsistent.}
    \label{fig:effect}
\end{figure}

In FL, our previous work FedLSR \cite{FedLSR} firstly points out that this effect can also exist in FL via empirical observations from experiments.
However, this previous observation can be incomplete since their experiments are based on two ideal assumptions: \textbf{(i)} Data distributions across clients are IID, which is an ideal assumption in FL \cite{noniidsurvey}; \textbf{(ii)} The noise rates across clients are the same, which is also an ideal assumption since clients usually adopt different labeling approaches as discussed in Section \ref{RW}.
Thus, this conclusion is actually incomplete to reflect how noisy labels affect FL.
To further observe this effect in the context of FL, we refer to more generalized settings on the synthetic noisy dataset CIFAR-10, where data across clients are of various noise rates.
We also follow the experimental settings in \cite{FedLSR} and illustrate the results in Figure \ref{fig:effect}(a), and the above-mentioned warm-up pattern can be basically observed under the two ideal assumptions.
Our experiments cover both IID and Non-IID data partitioning scenarios and we illustrate the results in Figure \ref{fig:effect}(b-c). 
Through test accuracy curves in Figure \ref{fig:effect}(b-c), we find this effect does not consistently hold when clients have various noise rates, and there is no so-called warm-up period under this more generalized setting. 
This effect can exist in centralized setting \cite{memorization}, but not absolutely holds in FL since noisy labels in FL have more complicated effects to training, which cannot be simply explained by memorization effect still exists in FL. Thus, we switch to another perspective to inspect how noisy labels affect FL.

\subsection{Dimensional Collapse}
\label{dimensionalCollapse}

\begin{figure}[tbp]
    \centering
    \includegraphics[width=\linewidth]{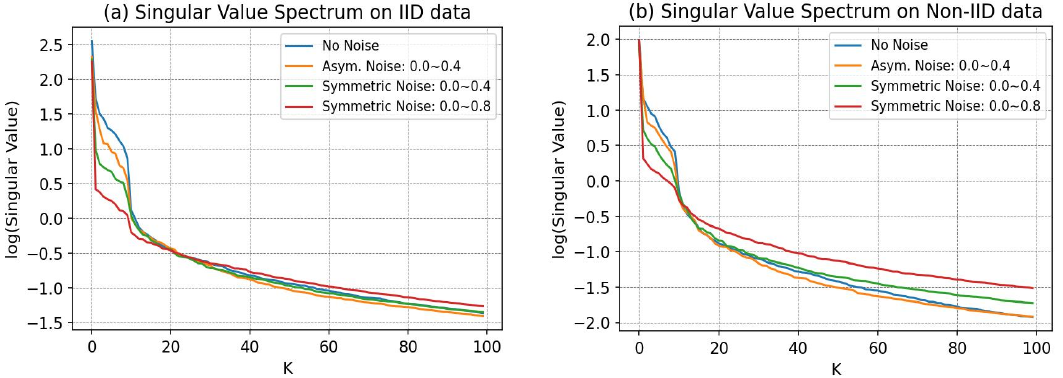}
    \caption{Noisy labels cause dimensional collapse on both (a) IID clients' data and (b) Non-IID clients' data. We plot singular values of the covariance matrix of representations in the descending order. X-axis is the index of singular values and the Y-axis is the logarithm of singular values. \textbf{We observe an evident scale decrease among the largest 20 singular values in X-axis.}}
    \label{fig:svd}
\end{figure}

To further inspect how noisy labels affect FL, we refer to dimensional collapse, which is recently discussed in the field of representation learning \cite{directCLR,decorr,sasvd}.
Classic DNN architecture for recognition is composed of a representation extractor and classifier \cite{moon,directCLR}, and the output representations reflect the underlying high-dimensional structure of data in representation space.
Since singular values of the covariance matrix of representations can provide a comprehensive characterization of the distribution of high dimensional representations \cite{decorr}, we study the output representations of test dataset by the final global model shown in Figure \ref{fig:svd}.
We rescale the singular values via logarithm following \cite{selfie}.
The result suggests larger noise rate causes the global model to suffer from more severe dimensional collapse, whereby representations are biased to residing in a lower-dimensional space.
Specifically, it is a form of oversimplification in terms of deep networks, where the representation space is not being fully utilized to discriminate diverse data of different classes \cite{decorr,directCLR}.
This also indicates that during the training process, the model extracts degraded representations before it generates inferior predictions in presence of distributed label noise. 

\begin{algorithm}[bp] 
    \caption{The computation process of $\mathcal{L}_{SVD}$}
    \label{algo_decorrloss}
    \LinesNumbered
    \KwIn{Representation matrix $\mathbf{M}$ of $\mathbf{B}$ samples}
    \KwOut{$\mathcal{L}_{SVD}$}
    $\epsilon \gets 10^{-8}$  ;  \textcolor{black}{// To avoid zero denominator in line 6.}  \\ 
    \textcolor{black}{/*Normalization*/}\\
    $\bar{\mathbf{M}} \gets \text{Mean}(\mathbf{M}, \text{dim}=0, \text{keepdim}=\text{True})$ \;
    $\mathbf{M} \gets \mathbf{M} - \bar{\mathbf{M}}$ \;
    $\sigma_{\mathbf{M}} \gets \text{Var}(\mathbf{M}, \text{dim}=0, \text{keepdim}=\text{True})$ \;
    $\mathbf{M} \gets \frac{\mathbf{M}}{\sqrt{\epsilon + \sigma_{\mathbf{M}}}}$ \;
    \textcolor{black}{/*Compute the correlation matrix*/}\\
    $\mathbf{corr} \gets \mathbf{M}^T \cdot \mathbf{M}$ \;
    \textcolor{black}{/* Extract the non-diagonal elements of $\mathbf{corr}$*/}\\
    $\hat{corr} \gets \text{remove\_diagonal\_elements} (\mathbf{corr})$ \;
    $\hat{corr} \gets Flatten(\hat{corr}) $  ;    \\ 
     $\mathcal{L}_{SVD} \gets (\hat{corr}^2.mean())/\mathbf{B}$  \;
    \Return $\mathcal{L}_{SVD}$ \;
\end{algorithm}

We further combine a regularization-based technique inspired by FedDecorr \cite{decorr} with existing methods to mitigate the degrading of representation by optimizing the output representation.
FedDecorr initially aims to solve the Non-IID issues in FL but we aim to exploit this regularization method to check whether this can help to improve robustness against label noise.
Detailedly, we apply a representation regularization term during local training that encourages different dimensions of representations to be uncorrelated from the perspective of singular value decomposition, and we name it \textit{SVD loss}.
We suppose $X$ denotes the training samples of each batch, and $(\lambda_1,\ldots,\lambda_d)$ are singular values of the covariance matrix of the representation matrix $M$ within each batch and $d$ denotes the number of singular values.
To mitigate the representation gap and enrich the representation space, the proposed \textit{SVD loss} aims to penalize the variance among the singular values,
while discouraging the tail singular values from collapsing to $0$.
This can be achieved by constraining the correlation matrix \cite{pca} $K_X$ of the representation matrix $M$ by the following below computation:
\begin{equation}
\mathcal{L}_{\text{SVD}} (X) = \frac 1 { d ^ 2 }\|K_X\|_{\text{F}}^2,    
\end{equation}

where $F$ denotes the Frobenius norm (F-norm \cite{fnorm}) of the correlation matrix $K_X$. This can be achieved by decreasing non-diagonal entries of $K_X$.
We provide pseudo codes in Algorithm \ref{algo_decorrloss} to showcase the computing process. More theoretical proof is available in \cite{decorr}.

Following the similar motivation to optimize the representation space, after we combine the above $L_{SVD}$ loss term with existing methods, we find the final performance for almost all methods gets improved, and we will show the experimental results and further discuss in Section \ref{sec:optRep}.

\section{Experiments}
\label{sec:exp}
\subsection{Implementation  Details}
\label{ExpSet}
All main experiments are implemented by Pytorch \cite{paszke2017automatic} on Nvidia$^\circledR$ RTX 3090 GPU.
To balance both precision and recall \cite{fedrn,selfie}, we mainly report the average F1 score \cite{f1score} of the global model on the clean testset $\mathcal{D}_{test}$ over last 10 rounds, which is a common metric \cite{co-teaching,FedLSR, RFL,cpf,fedelc}.
All experiments are averaged over 3 seeds for fair comparison.
Mixed precision training \cite{mixedPrecision} is utilized to accelerate training.
Experimental datasets are discussed in Section \ref{sec:datasets} and baselines are discussed in \ref{sec:baselines}.
For the image recognition task, we use the ResNet \cite{kaiming} as base models.
For the text classification task, we use FastText \cite{fasttext} as the base model, and the word embedding dimension is fixed to 256.
We generate $N=100$ clients, and 10 clients participate in FL in each communication round, which is a vanilla setting in FL \cite{fedavg,FedLSR}. SGD optimizer is selected with a learning rate of $0.01$, momentum of $0.9$ and weight decay of $5e^{-4}$ referring to \cite{fedelc,decorr}. The local epochs $e=5$ and the local training batch size is 64. The total communication round number is 120 for the image recognition task and 60 for the text classification task.

\subsection{Analysis on Datasets with Synthetic Noisy Labels}
\label{MainExp}

For the image recognition task, our main experiments are conducted on two benchmark datasets CIFAR-10 and CIFAR-100 with diverse Non-IID partitioning and synthetic label noise. 
The F1 scores of existing approaches on different non-IID data partitioning under various label noises are shown in Table \ref{tab:cifar10} and  Table \ref{tab:cifar100}. 
Through experimental results, we propose following observations:
\begin{itemize}[leftmargin=0.5cm]
    \item \textbf{Given conditions where clients' datasets are both Non-IID and noisy, no method consistently outperforms the others in all settings.} 
    Interestingly, general FL methods are relatively robust when noise rates across clients are relatively low (e.g. $0.0-0.4$).
    For NLL methods, Co-teaching \cite{co-teaching} and SELFIE \cite{selfie} are relatively more robust ones.
    For FNLL methods, FedNoRo \cite{FedNoRo} and FedELC \cite{fedelc} serve as relatively robust methods.
    The co-existence of both Non-IID and noisy characteristics of distributed data makes learning a robust model in the context of FL much harder.
    Therefore, there is still a need for a robust method to effectively handle noisy labels in FL while maintaining stable performance, which remains space for future works to explore. Perhaps future works can consider introducing suitable training framework to utilize peer networks \cite{co-teaching} in the context of FL.
    \item \textbf{Byzantine robust FL methods are less robust against distributed noisy clients.} 
    Byzantine robust FL methods are firstly proposed in traditional distributed learning \cite{limu,krum,TrimmedMean}, which are not tailored to consider the Non-IID characteristic of FL. 
    Among these methods, Krum \cite{krum} almost consistently performs worse than others.
    Since Krum cautiously select one local model from models from $S_t$ as the most reliable one as the updated global model, it does not fully utilize the underlying information provided by other clients which can also provide useful information.
    This reflects the importance of model aggregation in FL to aggregate knowledge with distributed clients' data to a certain extent.
    Meanwhile, robust aggregation methods are mainly discussed in byzantine robustness in FL, which supposes the ratio of bad participants of FL is relatively low.
    Actually, this ratio is hard to obtain considering the unknown noise levels of all clients in real-world practice.
    RFA \cite{rfa} is the most robust method for this group considering all scenarios with synthetic noisy labels, which can be considered in future works.
    \item \textbf{Apart from challenges of Non-IID and noisy data across clients, total class number $M$ is also a challenge.} 
    Compared with CIFAR-10, we find some methods evidently suffer from evident performance decrease on CIFAR-100 with synthetic label noise, which has much more classes $M$.
    Co-teaching \cite{co-teaching}, FedNoRo \cite{FedNoRo} and FedELC \cite{fedelc} are generally more robust than other methods on synthetic noisy CIFAR-100, and these methods follow the intuitive which aims to select more reliable samples or clients and give less weight to the possibly unreliable ones.
    Interestingly, general FL methods with simple designs are relatively more robust than many other methods considering all scenarios on synthetic noisy CIFAR-100. 
    For FNLL methods, FedLSR \cite{FedLSR} suffers on all scenarios on CIFAR-100. 
    This phenomena is also observed and the underlying reason is further investigated in its follow-up work \cite{lsg}, which claims the utilization of sharpening prediction technique should be more cautious when the total class number $M$ is large.
    RFL \cite{RFL} also suffers on most scenarios on synthetic noisy dataset CIFAR-100.
    We suppose that the underlying reason lies in the difficulty of fully capturing well-learned class centroids for each class on distributed Non-IID and noisy data for all 100 classes.
    For FedRN \cite{fedrn}, we reckon this method achieves inferior performance mainly due to the unreliable model pool located on the server side, which is the core implementation of FedRN.
    Considering all challenging characteristics, the server-side model pool possibly maintains various inferior models, which brings difficulties to obtain extra reliable and effective knowledge from other clients' models for each client.
    Therefore, further efforts are encouraged to propose a more robust method which can perform well on synthetic noisy datasets with higher total class number $M$.
\end{itemize}

\begin{table*}[t!]
\centering
\caption{F1 score\,(\%) on CIFAR-10 with three types of Non-IID partitioning and four synthetic label noise patterns. In each column, the \textbf{bold} denotes the best method across all groups while the \underline{underlined} denotes the best method across the belonged group.}
\label{tab:mainexp_cifar10}
\begin{tabular}{ccccccccccccc}\midrule\midrule
\textbf{Non-IID Type} &  \multicolumn{4}{c}{Shard\,($S=5$)} & \multicolumn{4}{c}{\textit{Dir.} \,($\beta=1.0$)} & \multicolumn{4}{c}{\textit{Dir} ($\beta=0.5$)}  \\\midrule
\textbf{Noise Type} & \multicolumn{2}{c}{Symmetric} & {Pairflip} & {Mixed} & \multicolumn{2}{c}{Symmetric} & {Pairflip} & {Mixed} & \multicolumn{2}{c}{Symmetric} & {Pairflip} & {Mixed} \\
\textbf{Noise Rate} & \!\!0.0--0.4\!\! & \!\!0.0--0.8\!\! & \!\!0.0--0.4\!\! & \!\!0.0--0.4\!\! & \!\!0.0--0.4\!\! & \!\!0.0--0.8\!\! & \!\!0.0--0.4\!\! & \!\!0.0--0.4\!\! & \!\!0.0--0.4\!\! & \!\!0.0--0.8\!\! & \!\!0.0--0.4\!\! & \!\!0.0--0.4\!\! \\\midrule
FedAvg \cite{fedavg}     & 60.09 & 38.66 & 54.33 & 59.70 & \underline{73.38} & \underline{54.15} & \underline{76.82} & \underline{75.54} & 62.69 & 44.47 & 63.83 & 66.94 \\

FedProx \cite{fedprox} & 55.92 & 37.94 & 54.32 & \underline{59.71} & 71.21 & 53.45 & 76.75 & 75.53 & \underline{63.34} & 43.18 & \underline{66.64} & 66.93 \\
FedExP \cite{FedExP}   & \underline{61.32} & \underline{38.92} & \underline{55.22} & 59.36   & 73.17 & 54.00 & 76.78 & 75.24 & 62.72 & \underline{44.58} & 63.47 & \underline{67.18} \\
\midrule

TrimmedMean \cite{TrimmedMean}    & 49.20 & 30.88 & 42.00 & 46.50 & 63.30 & 43.05 & 64.60 & 63.56 & 59.74 & 41.01 & 54.90 & 57.88 \\
Krum \cite{krum} & 20.33 & 15.31 & 13.77 & 12.39 & 34.10 & 25.81 & 38.21 & 44.46 & 27.59 & 19.56 & 15.32 & 19.79 \\
Median \cite{median}  & 56.39 & \underline{43.65} & 38.23 & 49.25 & 70.45 & \underline{56.03} & 72.30 & 71.27 & 63.40 & \underline{47.52} & 64.21 & \underline{65.37} \\
RFA \cite{rfa} & \underline{60.62} & 40.17 & \underline{54.27} & \underline{57.69} & \underline{71.63} & 54.32 & \underline{74.13} & \underline{73.66}  & \underline{64.34} & 46.10 & \underline{65.61} & 63.88 \\

\midrule
Co-teaching \cite{co-teaching}     & \textbf{\underline{64.83}} & 52.62 & 53.27 & 59.39 & 72.97 & 53.99 & \underline{75.69} & 72.69 & \underline{63.46} & 43.59 & \underline{62.63} & \underline{67.35} \\
Co-teaching+ \cite{co-teaching+} & 47.42 & 48.51 & 49.97 & 51.27 & 63.22 & 58.09 & 60.69 & 61.69 & 44.08 &42.39 & 35.48 & 41.12 \\
JointOptim \cite{jointopt} & 57.67 & 50.41 & 53.78 & 55.88 & 64.55 & 59.46 & 64.69 & 63.91 & 57.15 & 51.94 & 54.41 & 57.02 \\
SELFIE \cite{selfie}   & 63.77 & 49.77 & \underline{57.48}& \underline{60.01} & \underline{74.61} & 59.01& 73.58 & \underline{74.54} &63.14& 49.52 & 58.40 & 62.69 \\
Symmetric CE \cite{symmetricCE}  & 59.42 & \underline{54.61} & 56.31 & 57.00 & 72.74 & \textbf{\underline{67.66}} & 70.44 & 72.39 & 60.29 & \underline{53.24} & 56.90 & 56.81 \\  
DivideMix \cite{Dividemix} & 53.60 & 53.81 & 48.00 & 55.10 & 63.17 & 59.94 & 65.16 & 65.00 & 58.02 & 48.12 & 61.37 & 55.97 \\
\midrule
RFL \cite{RFL}      & 41.28 & 14.82 & 42.71 & 51.91& 61.73 & 43.94 & 61.67 & 60.42 & 46.57 & 19.28 & 43.76 & 43.59 \\
FedLSR \cite{FedLSR} & 57.93 &\textbf{ \underline{56.14}} & 53.41 & 54.99 & 70.58 & 66.54 & 70.81 & 67.74 & 54.42 & 53.39 &53.38& 48.84 \\
FedRN \cite{fedrn} & 50.30 & 42.89 & 43.82 & 43.70 & 67.37 & 58.45 & 66.23 & 59.76 & 49.74 & 48.20 & 44.53 & 47.61 \\
FedNoRo \cite{FedNoRo}& 56.99 & 41.50 & 58.10 & 64.86 & 73.26 & 66.98 & 77.73 & 74.88 & 71.33 & 57.34 & 72.39 &  72.33 \\
FedELC \cite{fedelc} & \underline{59.77} & 43.96 & \textbf{\underline{65.90}} & \textbf{\underline{70.02}} & \textbf{\underline{74.77}} & \underline{67.11} & \underline{\textbf{77.81}} & \textbf{\underline{75.90}} & \textbf{\underline{71.55}} & \textbf{\underline{57.67}} & \textbf{\underline{72.81}} & \textbf{\underline{72.99}}
\\ \bottomrule
\end{tabular}
\label{tab:cifar10}
\end{table*}

\begin{table*}[t!]
\caption{F1 score\,(\%) on CIFAR-100 with three types of Non-IID partitioning and four synthetic label noise patterns.}
\centering
\begin{tabular}{ccccccccccccc}\midrule\midrule
\textbf{Non-IID Type} &  \multicolumn{4}{c}{Shard\,($S=20$)} & \multicolumn{4}{c}{\textit{Dir.} \,($\beta=1.0$)} & \multicolumn{4}{c}{\textit{Dir} ($\beta=0.5$)}  \\\midrule
\textbf{Noise Type} & \multicolumn{2}{c}{Symmetric} & {Pairflip} & {Mixed} & \multicolumn{2}{c}{Symmetric} & {Pairflip} & {Mixed} & \multicolumn{2}{c}{Symmetric} & {Pairflip} & {Mixed} \\
\textbf{Noise Rate} & \!\!0.0--0.4\!\! & \!\!0.0--0.8\!\! & \!\!0.0--0.4\!\! & \!\!0.0--0.4\!\! & \!\!0.0--0.4\!\! & \!\!0.0--0.8\!\! & \!\!0.0--0.4\!\! & \!\!0.0--0.4\!\! & \!\!0.0--0.4\!\! & \!\!0.0--0.8\!\! & \!\!0.0--0.4\!\! & \!\!0.0--0.4\!\! \\\midrule
FedAvg \cite{fedavg}      & 32.33 & 20.45 & 34.48 & 35.23 & \underline{40.09} & \underline{27.23} & \underline{45.61} & \underline{41.88} & \underline{39.59} & \underline{26.07} & 43.43 & \underline{42.02} \\


FedProx \cite{fedprox} & 31.31 & 19.81 & 31.30 & 32.43 & 37.13 & 26.10 & 40.71 & 39.00 & 36.27 & 23.99 & 41.59 & 38.57 \\
FedExP \cite{FedExP}    & \underline{32.91} & \underline{20.60} & \underline{34.77} & \underline{35.42} & 39.41 & 27.04 & 44.69 & 41.40 & 39.47 & 25.78 & \underline{43.92} & 41.93 \\
\midrule

TrimmedMean \cite{TrimmedMean}    & 24.43 & 13.45 & 20.84 &  23.12 & 34.12 & 22.99 & 40.00 & 34.73 & 33.28 & 17.69 & 36.02 & 34.56 \\
Krum \cite{krum} & 1.45  & 1.21 & 1.54 & 1.31 & 13.74 & 10.44 & 15.92 & 14.98 & 8.20 & 4.29 & 7.59 & 7.06 \\
Median \cite{median}  & 11.03 & 5.99 & 8.39 & 7.61 & 33.83 & 21.38 & 36.93 & \underline{37.01} & 27.71 & 17.41 & 31.89 & 30.42 \\
RFA \cite{rfa} & \underline{33.03} & \underline{19.52} & \underline{34.88} & \underline{35.23} & \underline{40.93} & \underline{26.29} & \underline{44.33} & 44.65 & \underline{41.04} & \underline{25.32} & \underline{42.53} & \underline{41.38} \\

\midrule

Co-teaching \cite{co-teaching}   & \textbf{\underline{40.03}}& 26.20 & \underline{37.59} & 38.78 & \textbf{\underline{46.75}} & 30.74 & 46.51 & \textbf{\underline{46.79}} & \textbf{\underline{46.28}} & 30.70 & \textbf{\underline{46.13}} & \textbf{\underline{47.37}} \\
Co-teaching+ \cite{co-teaching} & 19.99 & 17.43 & 18.73 & 20.08 & 30.99 & 23.96 & 30.40 & 30.46 & 27.56 & 23.18 & 28.67 & 28.09 \\
JointOptim \cite{jointopt}  & 22.48 & 19.84 & 22.46 & 22.53 & 27.57 & 22.53 & 27.86 & 27.75 & 26.42 & 21.83 & 26.26 & 26.39 \\
SELFIE \cite{selfie}     & 39.91 & 23.68 & 39.43 & \textbf{\underline{39.21}} & 44.01 & 30.40 & \textbf{\underline{47.02}} & 46.24 & 43.37 & 29.10 & 44.70 & 45.04 \\
Symmetric CE \cite{symmetricCE}  & 29.18 & 26.43 & 29.40 & 30.59 & 38.35 & \underline{32.25} & 37.72 & 38.70 & 35.92 & \underline{32.06} & 35.43 & 36.71 \\ 
DivideMix \cite{Dividemix} & 32.88 & \textbf{\underline{28.65}} & 33.79 & 33.82 & 36.83 & 30.15 & 36.28 & 36.51 & 36.94 & 30.02 & 36.23 & 38.66 \\
\midrule

RFL \cite{RFL}     & 2.57 & 1.09 & 1.47 & 2.32 & 7.29 & 1.69 & 5.59 & 6.88 & 5.17 & 3.14 & 5.10 & 5.73 \\
FedLSR \cite{FedLSR}& 4.71 & 3.39 & 4.12 & 5.42 & 4.54 & 4.11 & 4.62 & 4.88 & 5.33 & 4.46 & 4.89 & 5.08 \\
FedRN \cite{fedrn} & 16.88 & 14.33 & 18.86 & 14.96 & 25.32 & 20.22 & 26.76 & 26.93 & 27.84 & 20.12 & 29.28 & 27.52 \\
FedNoRo \cite{FedNoRo} & \underline{35.96} & 21.85 & \textbf{\underline{40.11}} & 37.48 & \underline{42.61} & 32.77 & \underline{42.28} & \underline{45.03} & \underline{40.25} & 31.37 & 42.00 & 43.29 \\ 
FedELC \cite{fedelc} & 33.96 & \underline{23.54} & 37.61 & \underline{37.55} & 41.77 & \textbf{\underline{32.88}} & 41.92 & 44.47 & 39.40 & \textbf{\underline{32.11}} & \underline{42.33} & \underline{43.55} \\ \bottomrule
\end{tabular}
\label{tab:cifar100}
\end{table*}

\subsection{Investigation on Synthetic Noisy Text Classification}
Almost all NLL and FNLL works focus on image recognition tasks and we reckon it is far from enough for researches against noisy labels.
For the text classification task, in this study, we conduct experiments on the text classification dataset AGNews \cite{agnews} with synthetic noisy labels and Dirichlet partitioning ($\beta=1.0$). The F1 score results are shwon in Table \ref{tab:agnews}. 
We do not implement FedLSR \cite{FedLSR}, DivideMix \cite{Dividemix} and FedRN \cite{fedrn} since they are originally tailored for image related tasks and lack specific adaptions for the text classification task. 
Note that we find the performance of peer networks have bigger gaps (about 10\% to 20\%) for the text classification task, which may be attributed to the cumulative deviations and weight drifts \cite{moon,fedtrip} of these peer networks during the training process. We find Symmetric CE and RFA show overall best robustness considering all scenarios. We recommend future FNLL works can consider further improving robustness on this task.

\begin{table}[htbp]
\small
\caption{Evaluations on the  AGNews dataset for text classification.}
\centering
\small
\resizebox{0.9\linewidth}{!}{
\begin{tabular}{ccccc}
\toprule \toprule
\multirow{2}{*}{\textbf{Method}} & \multicolumn{4}{c}{\textbf{Label Noise Patterns}} \\
 & \begin{tabular}[c]{@{}c@{}}Symmetric\\ 0.0-0.4\end{tabular} & \begin{tabular}[c]{@{}c@{}}Symmetric\\ 0.0-0.8\end{tabular} & \begin{tabular}[c]{@{}c@{}}Pairflip\\ 0.0-0.4\end{tabular} & \begin{tabular}[c]{@{}c@{}}Mixed\\ 0.0-0.4\end{tabular} \\ \midrule
FedAvg \cite{fedavg} & 63.13& 58.77& 59.59&59.52\\
FedProx \cite{fedprox} & \underline{65.10}& \textbf{\underline{61.82}}& \underline{61.15}&\underline{63.66}\\
FedExP \cite{FedExP} & 62.73& 56.07 & 60.79 & 61.05\\ \midrule
TrimmedMean \cite{TrimmedMean} & 58.73& 51.72& 54.22&55.87\\
Krum \cite{krum} & 45.65 & 42.61& 44.33 &46.78\\
Median \cite{median} & 62.96& 60.72& 61.26&63.27\\
RFA \cite{rfa} & \underline{65.12}& \underline{61.43}& \underline{61.32}&\textbf{\underline{64.12}}\\ \midrule
Co-teaching \cite{co-teaching} &63.42 &58.67 &62.77 &62.11\\
Co-teaching+ \cite{co-teaching+} &61.07 &55.20 &58.45 &58.05\\
JointOptim \cite{jointopt} & 37.90 & 31.85 &36.34 & 38.48\\
SELFIE \cite{selfie} &37.85 &16.13 &27.35 &35.61\\
Symmetric CE \cite{symmetricCE} & \textbf{\underline{65.40}} & \underline{59.47} & \textbf{\underline{62.93}} & \underline{63.28} \\ 
DivideMix \cite{Dividemix} & -& -& -&-\\  \midrule
RFL \cite{RFL} &30.72 &28.07 &33.25 &29.95 \\
FedLSR \cite{FedLSR} & -& -& -&-\\
FedRN \cite{fedrn} & - & - & - &-\\
FedNoRo \cite{FedNoRo} & \underline{65.05} & \underline{59.77} & \underline{59.71} & \underline{62.86} \\ 
FedELC \cite{fedelc} & 61.13 & 57.89  & 57.51 & 60.12 \\ \bottomrule
\end{tabular}
}
\label{tab:agnews}
\end{table}

\subsection{Analysis on Datasets with Imperfect Human-annotation Errors}
We refer to CIFAR-10/100-N \cite{cifarn} to benchmark all methods under noisy datasets with real-world human annotation error patterns, which is rarely explored in previous FNLL works.
Therefore, we consider both three Non-IID data partitioning, as well as the IID data partitioning scenario.
The noisy labels are directly given by human annotators \cite{cifarn}. All results are listed in Table \ref{tab:cifar10N} and \ref{tab:cifar100N}.
For  CIFAR-10-N dataset, one interesting finding is that general FL methods consistently show higher robustness against most methods.
For NLL and FNLL methods, we find Co-teaching, FedNoRo and FedELC show higher performance against their counterparts.
This finding is following the finding on synthetic noisy datasets in Section \ref{MainExp}, which shows the potential of loss-based separation approach.
Moreover, fine-grained classwise loss-based clean client selection utilized in FedNoRo and FedELC shows higher performance than the coarse-grained class-agnostic loss-based local training method in Co-teaching (see Section \ref{RW}).
Regarding byzantine robust FL group, except RFA, most methods are all affected more by Non-IID partitioning, compared with the other three groups.
Krum almost shows the worst performance, which is in accordance with the finding on synthetic noisy datasets in Section \ref{MainExp}.
For CIFAR-100-N dataset which has more classes and more complex noise transition patterns (see Figure \ref{fig:cifarn})  than CIFAR-10-N, the same interesting finding can also be drawed.
General FL methods nearly show higher robustness against all other methods, which is also in accordance with the finding on the synthetic noisy CIFAR-100 dataset (see Section \ref{MainExp}).
We can also observe that many FNLL methods (RFL \cite{RFL}, FedLSR \cite{FedLSR}, and FedRN \cite{fedrn}) get struggling when the class number $M$ gets higher.
Overall, FedNoRo, FedELC and Co-teaching have higher robustness against their counterparts.

Given above findings, we suggest future studies can focus on improving loss-separation based methods like FedNoRo, FedELC and Co-teaching.
Moreover, the general FL methods interestingly show higher robustness against human-annotation error patterns, which reflects simple methods can also have the potential to achieve robustness against label noise.

\begin{table}[htbp]
\caption{Evaluation results (metric: F1 score) on CIFAR-10-N which approximated human annotation error patterns.}
\label{tab:cifar10N}
\resizebox{\linewidth}{!}{%
\begin{tabular}{ccccc}
\toprule \toprule
\multirow{3}{*}{\textbf{Method}} & \multicolumn{4}{c}{\textbf{Data Partitioning}} \\ \cmidrule{2-5} 
 & \textbf{IID} & \multicolumn{3}{c}{\textbf{Non-IID}} \\
 & - & Sharding & \textit{Dir} ($\beta=1.0$) & \textit{Dir} ($\beta=0.5$) \\ \midrule
FedAvg \cite{fedavg}\ & \textbf{\underline{88.38}}& 64.10 & 83.42 & 72.96 \\
FedProx \cite{fedprox} & 87.68 & \underline{68.83} & \underline{\textbf{84.90}} & \underline{79.23} \\
FedExp \cite{FedExP} & 88.23 & 64.02 & 83.66 & 73.09 \\ \midrule
TrimmedMean \cite{TrimmedMean} & 85.79 & 47.64 & 77.66 & 64.04 \\
Krum \cite{krum} & 75.55 & 15.12 & 59.23 & 29.94 \\
Median \cite{median} & 85.94 & 33.15 & 81.34 & 71.09 \\
RFA \cite{rfa} & \underline{88.24} & \underline{62.94} & \underline{84.09} & \underline{75.71} \\ \midrule
Co-teaching \cite{co-teaching} & \underline{84.38} & \underline{63.75} & \underline{76.56} & 59.71 \\
Co-teaching+ \cite{co-teaching+} & 78.29 & 48.59 & 63.45 & 47.60 \\
JointOptim \cite{jointopt} & 78.22 & 56.97 & 67.90 & 59.51 \\
SELFIE \cite{selfie} & 83.43 & 58.34 & 73.77 & 55.12 \\
Symmetric CE \cite{symmetricCE} & 83.53 & 59.56 & 75.04 & \underline{60.44} \\
DivideMix \cite{Dividemix} & 75.72 & 48.69 & 65.55 & 56.46 \\ \midrule
RFL \cite{RFL} & 75.37 & 50.07 & 64.00 & 52.29 \\
FedLSR \cite{FedLSR} & 81.79 & 48.35 & 73.85 & 49.46 \\
FedRN \cite{fedrn} & 75.61 & 48.52 & 70.14 & 46.72 \\
FedNoRo \cite{FedNoRo} & 87.08 & 77.14 & 84.04 & 81.41\\
FedELC \cite{fedelc} &  \underline{87.58} & \underline{\textbf{78.10}}  & \underline{84.31}  & \underline{\textbf{82.30}}\\ \bottomrule
\end{tabular}
}
\end{table}

\begin{table}[htbp]
\caption{Evaluations on CIFAR-100-N with human annotation errors.}
\label{tab:cifar100N}
\resizebox{\linewidth}{!}{%
\begin{tabular}{ccccc}
\toprule \toprule
\multirow{3}{*}{\textbf{Method}} & \multicolumn{4}{c}{\textbf{Data Partitioning}} \\ \cmidrule{2-5} 
 & \textbf{IID} & \multicolumn{3}{c}{\textbf{Non-IID}} \\
 & - & Sharding & \textit{Dir} ($\beta=1.0$) & \textit{Dir} ($\beta=0.5$) \\ \midrule
FedAvg \cite{fedavg} & {\ul \textbf{57.16}} & 44.22 & {\ul \textbf{56.69}} & 55.56 \\
FedProx \cite{fedprox} & 54.47 & 44.46 & 54.08 & 54.40 \\
FedExp \cite{FedExP} & 56.97 & {\ul 44.97} & 55.79 & {\ul \textbf{56.48}} \\ \midrule
TrimmedMean \cite{TrimmedMean} & 51.80 & 26.84 & 48.66 & 47.22 \\
Krum \cite{krum} & 34.43 & 2.03 & 23.16 & 11.30 \\
Median \cite{median} & 50.66 & 33.56 & 47.53 & 43.72 \\
RFA \cite{rfa} & \underline{57.08} & \underline{44.37} & \underline{55.17} & \underline{56.21} \\ \midrule
Co-teaching \cite{co-teaching} & {\ul 55.36} & {\ul 43.57} & {\ul 54.93} & 53.01 \\
Co-teaching+ \cite{co-teaching+} & 36.77 & 23.49 & 35.79 & 31.16 \\
JointOptim \cite{jointopt} & 33.27 & 24.18 & 31.84 & 30.33 \\
SELFIE \cite{selfie} & 54.67 & 43.51 & 54.79 & {\ul 53.12} \\
Symmetric CE \cite{symmetricCE} & 44.23 & 31.86 & 42.16 & 36.97 \\
DivideMix \cite{Dividemix} & 40.92 & 37.54 & 42.93 & 41.93 \\ \midrule
RFL \cite{RFL} & 16.78 & 3.10 & 14.63 & 10.27 \\
FedLSR \cite{FedLSR} & 7.08 & 5.59 & 6.05 & 7.10 \\
FedRN \cite{fedrn} & 33.12 & 21.01 & 30.38 & 27.39 \\
FedNoRo \cite{FedNoRo} & 55.59 & 52.70 & 52.38 & 49.54 \\
FedELC \cite{fedelc} &  \underline{55.87} & \underline{\textbf{53.07}}  & \underline{52.99}  & \underline{53.02} \\ \bottomrule
\end{tabular}
}
\end{table}

\subsection{Analysis on Dataset with Systematic Noisy Labels}
\label{clo1m}

Real-world evaluations are conducted over Clothing1M dataset \cite{clothing1m}. It is a large-scale dataset with $M$ = 14 categories, which contains 1 million clothing images crawled from several online shopping websites. It is reported that its overall noise rate is approximately 39.46\% with natural unstructured noise \cite{40}.
All images are resized to 224 × 224 and normalized.
Pretrained ResNet-50 \cite{kaiming} on ImageNet \cite{deng2009imagenet} is selected as the base model.
Other training details are in line with our previous works \cite{FedLSR,fedelc}.
In each communication round, 5 clients are selected to participate in each communication round of FL (40 rounds in total).
We report the average value of best test accuracy over three runs.
Related results are shown in Figure \ref{tab:testonclothing}.
From these results, we can observe that RFL \cite{RFL} shows higher accuracy, which indicates the extra global supervision can be effective for systematic unstructured label noise.

\begin{table}[htbp]
\caption{Evaluation results on large-scale systematic noisy dataset Clothing1M with random partitioning.}
\label{tab:testonclothing}
\centering
\begin{tabular}{cc}
\midrule
\midrule
 \textbf{Method}                 & \textbf{Best Test Accuracy} (\%)\\ \midrule
 FedAvg \cite{fedavg}                & 69.26\\ 
 FedProx \cite{fedprox}                & 67.62\\ 
 FedExP \cite{FedExP}      & 69.53 \\ \midrule
 TrimmedMean \cite{TrimmedMean}                        & 68.57\\ 
 Krum \cite{krum}      & 67.32 \\ 
 Median \cite{median}                   & 69.00 \\ 
 RFA \cite{rfa}                   & 69.23 \\ \midrule
 Co-teaching \cite{co-teaching} & 69.78 \\ 
 Co-teaching+ \cite{co-teaching+}      & 68.07 \\ 
 JointOptim \cite{jointopt}      & 70.51\\ 
 SELFIE \cite{selfie}                  & 69.97\\ 
 Symmetric CE \cite{symmetricCE}                 & 69.61 \\ 
 DivideMix \cite{Dividemix}                        & 70.08\\ \midrule
 RFL \cite{RFL}      & \textbf{71.77}\\ 
 FedLSR \cite{FedLSR}                        & 69.76\\ 
 FedRN \cite{fedrn}      & 68.60 \\ 
 FedCorr \cite{fedcorr}      & 69.92 \\ 
 FedNoRo \cite{FedNoRo}      & 70.54 \\ 
 FedELC \cite{fedelc}      & 71.64 \\ \bottomrule
\end{tabular}
\end{table}

\subsection{Batchsize Selection}
Batch size is a fundamental hyperparameter in federated learning (FL) systems, yet its impact on performance is often overlooked in existing works. Many FL methods \cite{Fednoisy, qinbinCrossSilo}, tend to use larger batch sizes, primarily to improve training efficiency. In our study, we evaluate the effect of different local batch sizes (32, 64, 128) on performance, using ResNet-18 as the base model. The results, presented in the left half of Table \ref{tab:ablation}, reveal that smaller batch size consistently yields better performance in most cases under limited communication rounds.
This performance variation can be attributed to the interplay between local training dynamics and the global aggregation process in FL. Specifically, with a limited communication budget, larger batch sizes reduce the number of local training iterations per round. As a result, the locally trained models are less refined and farther from their respective local optima before aggregation, leading to suboptimal global updates. Smaller batch sizes, on the other hand, allow for more frequent gradient updates, which better capture the local data distribution and improve the quality of the aggregated global model.
However, there is a trade-off: Smaller batch sizes increase the number of local iterations, which can lead to higher wall-clock training time and computational overhead on resource-constrained devices (e.g., mobile or IoT devices). Thus, while batch size = 32 achieves a good balance of performance and efficiency in our experiments, the optimal batch size in real-world settings may depend on the hardware constraints, available training time, and communication budget. Given sufficient resources, smaller batch sizes might be preferable for achieving better performance in FL systems.

\begin{table}[htbp]
\small
\caption{Ablation study (metric: F1 score) towards batchsize and base models on CIFAR-10 with Non-IID partitioning (dirichlet $\beta=1.0$) under symmetric noise (0.0-0.4).  R18, R34 and R50 are acronyms for the selction of ResNet-18/34/50.}
\small
\resizebox{\linewidth}{!}{
\begin{tabular}{ccccccc}
\toprule \toprule
\multirow{2}{*}{\textbf{Method}} & \multicolumn{3}{c}{\textbf{Batchsize}} & \multicolumn{3}{c}{\textbf{Model}} \\
 & 32 & 64 & 128 & R18. & R34. & R50. \\ \midrule
FedAvg \cite{fedavg} & \textbf{79.51} & 73.43 & 64.34 & 73.43 & 76.87 & 71.34 \\
FedProx \cite{fedprox} & \textbf{75.16} & 71.27 & 64.59 & 71.27 & 75.33 & 71.84 \\
FedExP \cite{FedExP} & \textbf{79.30} & 73.19 & 65.70 & 73.19 & 76.20 & 72.90 \\ \midrule
TrimmedMean \cite{TrimmedMean} & \textbf{69.09} & 63.35 & 57.92 & 63.35 & 68.03 & 60.83 \\
Krum \cite{krum} & \textbf{48.32} & 25.89 & 29.08 & 25.89 & 34.05 & 27.45 \\
Median \cite{median} & \textbf{73.46} & 70.45 & 64.33 & 70.45 & 72.21 & 66.40 \\
RFA \cite{rfa} & \textbf{74.55} & 71.63 & 65.87 & 71.63 & 73.01 & 67.22 \\ \midrule
Co-teaching \cite{co-teaching} & \textbf{79.73} & 72.95 & 66.30 & 72.95 & 76.90 & 74.25 \\
Co-teaching+ \cite{co-teaching+} & 60.33 & \textbf{63.27} & 57.64 & 63.27 & 57.34 & 51.92 \\
JointOptim \cite{jointopt} & 63.23 & \textbf{64.65} & 60.47 & 64.65 & 62.39 & 53.52 \\
SELFIE \cite{selfie} & \textbf{78.54} & 74.68 & 64.80 & 74.68 & 74.28 & 70.64 \\
Symmetric CE \cite{symmetricCE} & \textbf{73.71} & 72.76 & 64.84 & 72.76 & 67.66 & 69.01 \\
DivideMix \cite{Dividemix} & \textbf{65.35} & 63.19 & 60.51 & 63.19 & 63.52 & 55.16 \\ \midrule
RFL \cite{RFL} & \textbf{62.74} & 61.76 & 52.10 & 61.76 & 56.93 & 48.19 \\
FedLSR \cite{FedLSR} & \textbf{71.38} & 70.68 & 55.26 & 70.68 & 51.15 & 31.94 \\
FedRN \cite{fedrn} & \textbf{76.18} & 67.46 & 52.73 & 67.46 & 64.23 & 42.06 \\
FedNoRo \cite{FedNoRo} & \textbf{78.30} & 73.26 & 69.06 & 73.26 & 75.70 & 74.31 \\ 
FedELC \cite{fedelc} & \textbf{77.87} & 74.11 & 69.87 & 74.77 & 75.75 & 75.11 \\ \bottomrule
\end{tabular}
}
\label{tab:ablation}
\end{table}

\subsection{Base Model Selection}
The selection of base models plays a crucial role in deploying a FL system. Larger neural networks (NNs) typically have more parameters, which can improve their generalization ability when the data has correct labels \cite{overfit}. 
However, NNs can also overfit noisy training data \cite{overfit,memorization}.
This motivates us to study how different NNs perform in noisy conditions and whether larger models are more prone to overfitting noisy data or more robust to label noise in FL.
We fix the local batch size at 64 and extend the communication rounds to 160. Empirical results are presented in the right half of Table \ref{tab:ablation}.
Our observations include:
\textbf{(i)} A larger model does not necessarily offer greater robustness to noisy labels. For all methods, with limited communication budgets, ResNet-50 performs worse than ResNet-18 and ResNet-34. For NLL and FNLL methods, smaller NNs might be a better choice.
\textbf{(ii)} General FL methods are relatively robust to the choice of base model, while FNLL methods are more sensitive to base model selection. Notably, the FedLSR method experiences a significant performance drop with ResNet-50. This may be due to the unsuitable use of the sharpening operation (see Section \ref{RW} and \cite{FedLSR}), as analyzed in \cite{lsg}, and the fact that larger models are harder to fully optimize.
Based on our findings, we emphasize the importance of studying base model selection in future work. This is a key factor in achieving robust performance against noisy labels. Some recommended recent studies also provide their insights into the memorization effect in FL \cite{fedelr,flr}.

\subsection{Optimizing Deep Representation Space}
\label{sec:optRep}
\begin{figure}[htbp]
    \centering
    \includegraphics[width=\linewidth]{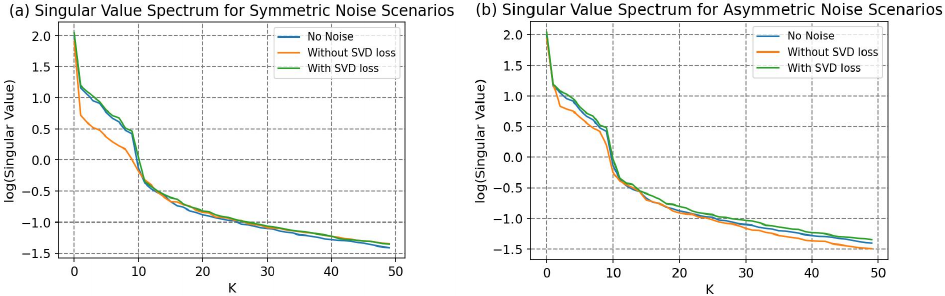}
    \caption{Singular values of the covariance matrix of representations in descending order for two noise types. X-axis (K) is the index of singular values and the Y-axis is the logarithm of the singular values. The No-noise line denotes training data are of no label noise.}
    \label{fig:svdloss}
\end{figure}

Herein we introduce an explicit technique \cite{decorr} on regularizing the representation space to study whether this technique will help to tackle the negative effect and dimensional collapse effect caused by noisy labels, which is discussed in Section \ref{dimensionalCollapse}. We list related experimental results in Table \ref{tab:svdloss}. 
From these results, we observe that we can generally achieve better robustness against label noise by combining most of the existing methods with the proposed \textit{SVD loss}, which enriches the representation space.
For FedELC, it takes relatively more complicated optimizing framework and multiple loss terms which also aims to learn the ground-truth labels, the performance shows little yet acceptable degrade after adding our proposed representation-aware technique. 
Taking FedAvg \cite{fedavg} as the example, we plot the singular values of the covariance matrix of representations (using the test dataset) in Figure \ref{fig:svdloss} and we find after applying this regularizing term, the dimensional collapse gap (discussed in Section \ref{dimensionalCollapse}) gets evidently mitigated. Through these exploration experiments, we find for future works, one potential approach to tackling noisy labels is to improve the learned representations.

\begin{table}[htbp]
\caption{Exploration experiments (metric: F1 score) on the potential of incorporating a representation regularization method.}
\label{tab:svdloss}
\begin{tabular}{ccccc}
\midrule \midrule
\multicolumn{1}{c}{\multirow{2}{*}{\textbf{Method}}} & \multicolumn{2}{c}{Symmetric (0.0-0.4)} & \multicolumn{2}{c}{Pairflip (0.0-0.4)} \\ \cmidrule{2-5} 
\multicolumn{1}{c}{} & - & + \textit{SVD loss} & - & + \textit{SVD loss} \\ \midrule
FedAvg \cite{fedavg} & 73.38 & 75.69 (+2.31) & 76.82 & 78.02 (+1.20) \\
FedProx \cite{fedprox} & 71.21 & 71.53 (+0.32) & 76.75 & 76.78 (+0.03) \\
FedExp \cite{FedExP} & 73.17 & 75.83 (+2.66) & 76.78 & 78.93 (+2.15) \\ \midrule
TrimmedMean \cite{TrimmedMean} & 63.30 & 65.10 (+1.80) & 64.60 & 66.68 (+2.08) \\
Krum \cite{krum}& 34.10 & 43.59 (+9.49) & 38.21 & 40.28 (+2.07) \\
Median \cite{median} & 70.37 & 71.72 (+1.35) & 72.30 & 73.38 (+1.08) \\
RFA \cite{rfa} & 71.63  &  72.12 (+0.49) &  74.13 & 74.48 (+0.35) \\ \midrule
Co-teaching \cite{co-teaching}& 72.93 & 77.78 (+4.85) & 75.67 & 77.75 (+2.08) \\
Co-teaching+ \cite{co-teaching+}& 63.19 & 67.00 (+3.81) & 60.67 & 65.91 (+5.24) \\
JointOptim \cite{jointopt}& 64.55 & 64.85 (+0.30) & 64.69 & 65.14 (+0.45) \\
SELFIE \cite{selfie}& 74.61 & 75.64 (+1.03) & 73.58 & 74.73 (+1.15) \\
Symmetric CE \cite{symmetricCE}& 72.74 & 74.04 (+1.30) & 70.44 & 71.26 (+0.82) \\
DivideMix \cite{Dividemix}& 63.17 & 63.81 (+0.64) & 65.16 & 66.01 (+0.85) \\ \midrule
RFL \cite{RFL}& 61.73 & 62.82 (+1.09) & 61.67 & 62.97 (+1.30) \\
FedLSR \cite{FedLSR}& 70.58 & 70.83 (+0.25) & 70.81 & 71.37 (+0.56) \\
FedRN \cite{fedrn}& 67.37 & 68.21 (+0.84) & 66.23 & 66.54 (+0.31) \\
FedNoRo \cite{FedNoRo}& 73.26 & 73.40 (+0.17) & 77.73 & 78.71 (+0.98) \\ 
FedELC \cite{fedelc}& 74.20 & 74.17 (-0.03) & 77.93 &  77.86 (-0.07) \\  \bottomrule
\end{tabular}
\end{table}

\subsection{Overhead Analysis}
We intuitively show the computation and communication cost analysis inspired by previous work \cite{fedrn}. Meanwhile, we also conduct experiments and plot on-device computation cost statistics per global communication round consumed by a client running these methods, relative to when it runs the vanilla FedAvg. We can observe the computation of DivideMix is much higher than others, since it maintains two peer networks, performs multiple forward computations and requires multiple data augmentation and mixup \cite{mixup}.
We also analyze the communication and computation cost of implemented baselines, as summarized in Table \ref{tab:summary_data}. Note that backward computation is more time-consuming than forward computation. Extra communication is compared with the vanilla FedAvg, which only transmits single model parameters between clients and servers. Notably, the evaluation statistics are collected based on our experimental settings with adequate GPU memory, and different experimental settings and specific implementation details may lead to certain differences. It reveals it is also important to derive a computation-friendly yet effective method to tackle noisy labels in FL. 

\begin{table}[htbp]
\caption{Analysis of communication (Comm.) and computation (Comp.) costs.  $\textbf{F}$ and $\textbf{B}$ are the computational cost of forward and backward computation of FedAvg; $e$ is local epochs in each communication round; $k$ is  neighbor number of FedRN. $\dagger$ denotes requiring extra data augmentation.}
\label{tab:summary_data}
\centering
\resizebox{\linewidth}{!}{%
\begin{tabular}{ccccc}
\midrule
\midrule
\textbf{Method} & \textbf{Extra Comm.}   & \textbf{Forward Comp.}    & \textbf{Backward Comp.}        \\ \midrule
FedAvg \cite{fedavg} & $\times$         & e\textbf{F}            & e\textbf{B}              \\ 
FedProx \cite{fedprox} & $\times$          & e\textbf{F}            & e\textbf{B}                    \\
FedExP \cite{FedExP} & $\times$          & e\textbf{F}            & e\textbf{B}                       \\ \midrule
TrimmedMean \cite{TrimmedMean} & $\times$         & e\textbf{F}            & e\textbf{B}              \\ 
Krum \cite{krum} & $\times$          & e\textbf{F}            & e\textbf{B}                   \\
Median \cite{median} & $\times$         & e\textbf{F}            & e\textbf{B}                     \\
RFA \cite{rfa} & $\times$         & e\textbf{F}            & e\textbf{B}                     \\ \midrule
Co-teaching  \cite{co-teaching}  & $\checkmark$          & 2e\textbf{F}            & 2e\textbf{B}              \\ 
Co-teaching+  \cite{co-teaching+}  & $\checkmark$          & 2e\textbf{F}            & 2e\textbf{B}                   \\
JointOptim \cite{jointopt} & $\times$          & e\textbf{F}            & e\textbf{B}              \\ 
SELFIE \cite{selfie} & $\times$          & e\textbf{F}            & e\textbf{B}                    \\
Symmetric CE \cite{symmetricCE} & $\times$          & e\textbf{F}            & e\textbf{B}                   \\
DivideMix \cite{Dividemix} & $\checkmark$          & 2e\textbf{F} (+ $\dagger$)         & 2e\textbf{B}                       \\ \midrule
RFL \cite{RFL} & $\checkmark$          & 2e\textbf{F}            & e\textbf{B}              \\ 
FedLSR \cite{FedLSR} & $\times$          & 2e\textbf{F} (+ $\dagger$)           & e\textbf{B}                   \\
FedRN \cite{fedrn}  & $\checkmark$         & (e+2k+1)\textbf{F}            & (k+e)\textbf{B}                   \\ 
FedNoRo \cite{FedNoRo}  & $\checkmark$         & (e+1)\textbf{F}            & e\textbf{B}                   \\ 
FedELC \cite{fedelc}  & $\checkmark$         & (e+1)\textbf{F}            & e\textbf{B}                   \\ 
\bottomrule
\end{tabular}%
}
\end{table}


\section{Discussion}
\label{sec:discuss}

We discuss the limitations of this work and propose some underlying directions to be explored in future works.

\textbf{Limitations.} We consider limitations of this study include but are not limited to:
\textbf{(i)} Apart from our conducted ablation study, there are still open problems that need to be addressed to complete this work. For example, how does the client participation rate (i.e. $\frac{|S_t|}{N}$ in Section \ref{prelin}) affect the final performance, and whether a higher participation rate implies more robust model performance? 
\textbf{(ii)} Since fine-grained hyper-parameter searching is quite time-consuming, in this work, we mainly follow the original papers for hyper-parameter selection. For some methods like co-teaching, we select mild hyper-parameters considering all diverse evaluated scenarios. With more careful hyper-parameter tuning, some  methods can achieve higher performance and they can be also considered for practical applications.
\textbf{(iii)} Noisy data are more than data with noisy labels, since original data can be noise-injected \cite{qinbinCrossSilo} or partially corrupted \cite{corrupted} such as low-quality images and speech recognition over noisy environments. Therefore, further investigation is necessary to explore other tasks with task-specific models in the context of FL. 

\textbf{Future works.} We propose three-fold future works:
\begin{itemize}[leftmargin=0.3cm] 
    \item \textbf{More real-world evaluation.}  All evaluations in this study are conducted on a whole clean test dataset which is an ideal assumption. Therefore, more real-world evaluation approaches and metrics \cite{quant} should be explored. For example, a server can consider taking the expense to collect a small-scale task-specific curated test dataset like \cite{rhfl}. Meanwhile, clients may evaluate other local models of other clients via a cross-validate approach \cite{clc,tdsc,chen2020focus}. 
    \item \textbf{Exploration on diverse tasks.} Since FL is an emerging research direction, it is hard to incorporate all well-noted methods in this study. Our study evaluates the image recognition task and text classification task in presence of noisy labels. However, in FL, noisy labels also exist in other tasks like graph learning \cite{fedrgl}, image segmentation \cite{feda3i} , recommendation system \cite{noisyFeedback}, and sequence learning \cite{feddshar}. Future works can explore these less explored tasks.
    \item \textbf{Simpler yet effective methods.} 
    Many recent works incorporate multiple existing techniques altogether to achieve relatively state-of-the-art performance, which correspondingly introduce multiple hyper-parameters to tune and are hard to be deployed in practice. 
    As analyzed in Section \ref{MainExp}, there still lacks effective methods to tackle both Non-IID and noisy data of more classes. 
    Nevertheless, some foundamental techniques (e.g. data augmentation \cite{augdesc}, data sampling \cite{anran,fednoil,smote-variants}, common regularization technique \cite{unleashing}, gradient clipping \cite{peijian}, generating  balanced and reliable instances \cite{fd} and self-supervised learning (discussed in Section \ref{dimensionalCollapse}) \cite{surveyonssl, Simclr}) should be revisited, and opportunities can exist in proposing simpler yet effective methods.
    Another meaningful trend lies in the possibility of correcting possible noisy labels within data \cite{fedelc,clc,dualoptim}.
\end{itemize}

\section{Conclusion}
\label{sec:conclusion}
In this study, we conduct a careful literature review on previous work related to tacking noisy labels in FL. 
We provide some insights to inspect why noisy labels impair FL from two perspectives, including deep network memorization effect and dimensional collapse.  
To facilitate future studies, we implement a unified and easy-to-use evaluation framework that supports eighteen methods on five image recognition datasets and one text classification dataset, and conduct extensive experiments to provide empirical evaluations of these methods. 
Based on our observations that learned representations show the dimensional collapse, we exploit a representation-aware regularization method and combine it with existing methods to achieve higher robustness against label noise.  
Finally, we discuss limitations of this work and provide three-fold directions for future works. 
We believe this work serves as the first comprehensive benchmark study towards federated learning against noisy labels, and it will have a broader impact on performing FL under weakly supervised conditions like medical analysis and network security, where FL has promising potential for making positive impacts on society.

\section*{Acknowledgments}
We appreciate our reviewers, our editors and our adminstrator for their efforts and expertise.
This work is supported by the National Natural Science Foundation of China (No. 62072436), the National Key Research and Development Program of China (2021YFB2900102) and Zhongguancun Lab.

\bibliographystyle{IEEEtran}
\bibliography{main}


\begin{IEEEbiography}[{\includegraphics[width=1in,height=1.25in,clip,keepaspectratio]{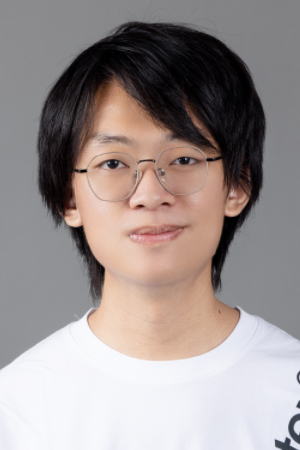}}]{Xuefeng Jiang}
is currently a Ph.D. candidate
with the Institute of Computing Technology, Chinese Academy of Sciences. Before that, he received his bachelor's degree with honors in Beijing
University of Posts and Telecommunications. His
research interests include distributed optimization and machine learning.
\end{IEEEbiography}

\begin{IEEEbiography}[{\includegraphics[width=1in,height=1.25in,clip,keepaspectratio]{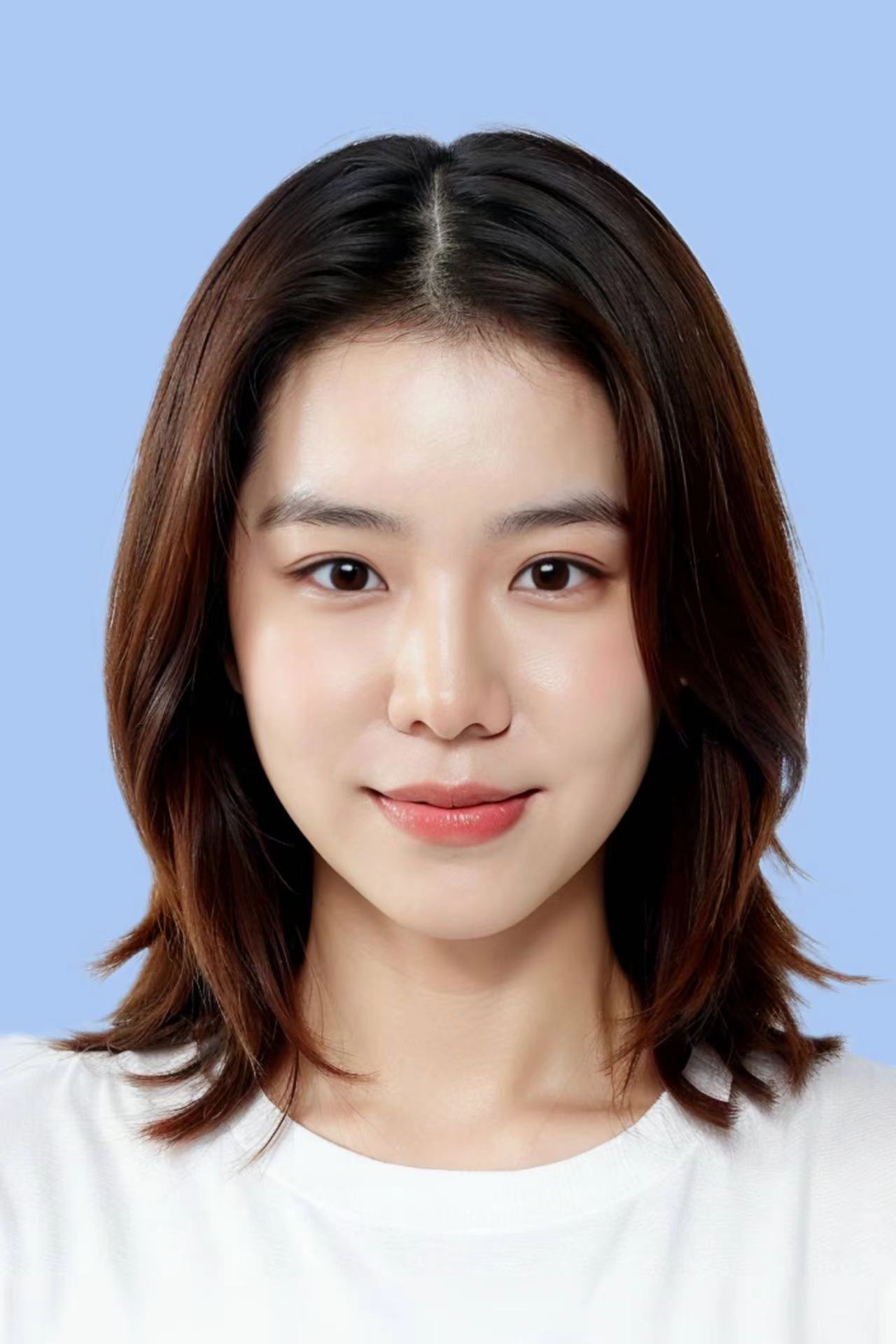}}]{Jia Li}
receives the master degree
with the Institute of Information Engineering, Chinese Academy of Sciences. Her
research interests include causal inference, trustworthy AI and generative AI.
\end{IEEEbiography}

\begin{IEEEbiography}[{\includegraphics[width=1in,height=1.25in,clip,keepaspectratio]{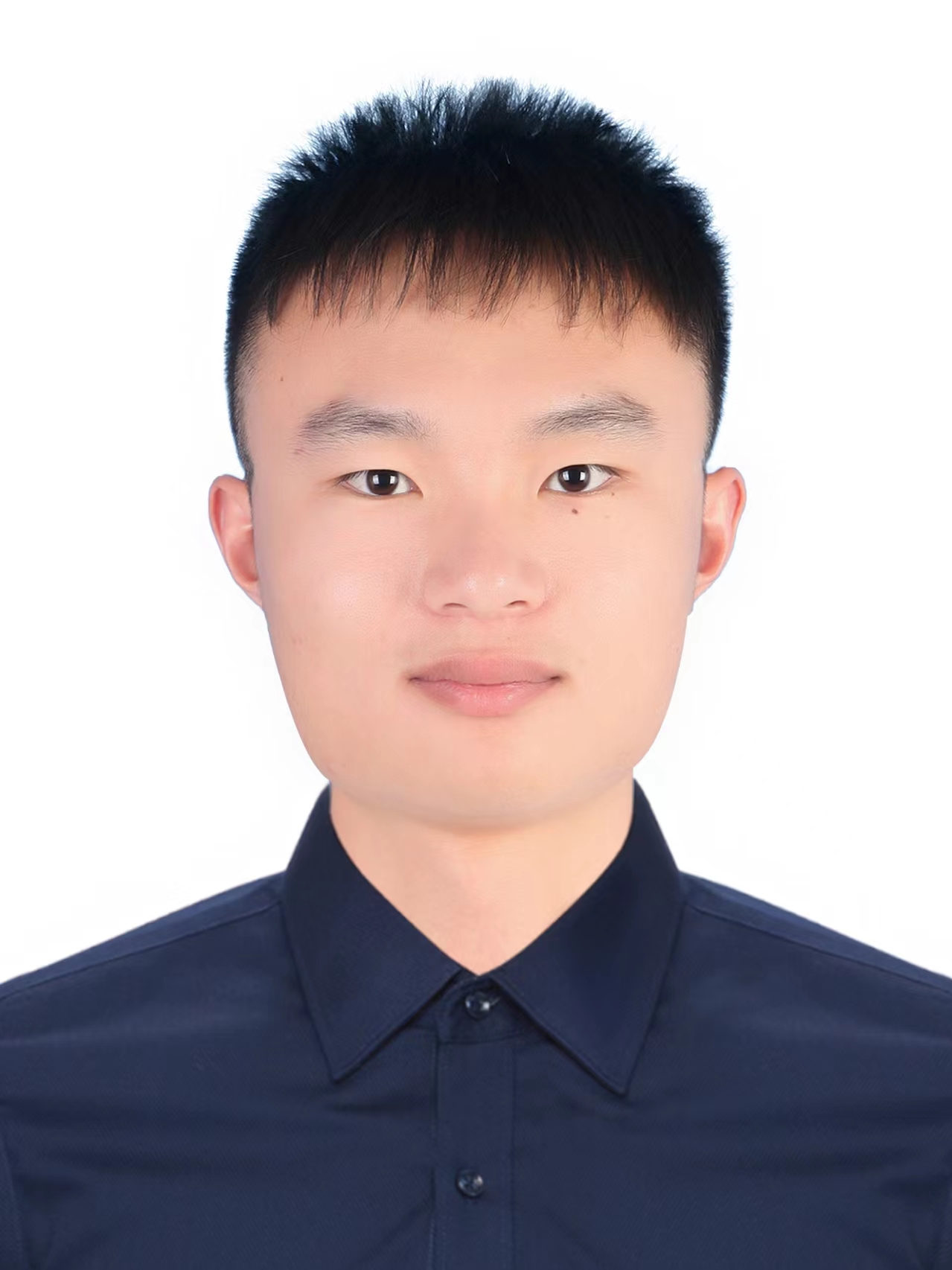}}]{Nannan Wu} is currently pursuing a Ph.D. at the Huazhong University of Science and Technology. Prior to embarking on his doctoral studies, he received his Bachelor's degree from the same institution in 2022. His research primarily focuses on federated learning and medical image analysis.
\end{IEEEbiography}

\begin{IEEEbiography}[{\includegraphics[width=1in,height=1.25in,clip,keepaspectratio]{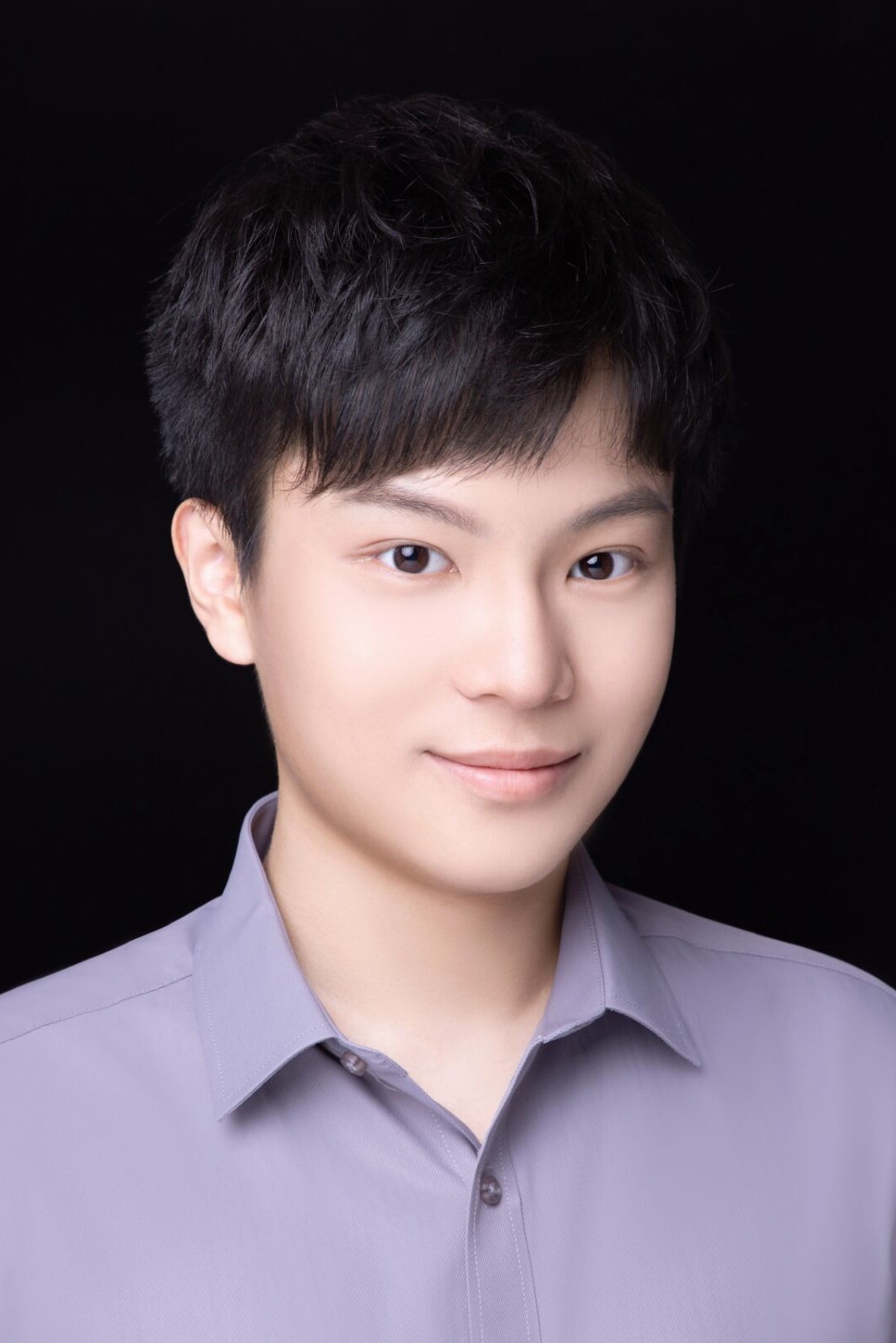}}]{Zhiyuan Wu}
(Member, IEEE) is currently a master student with the Institute of Computing Technology, Chinese Academy of Sciences. He has contributed several technical
papers to top-tier venues including IEEE Transactions on Parallel and Distributed Systems (TPDS), IEEE Transactions on Mobile Computing (TMC), and IEEE International Conference on Computer Communications (INFOCOM). His research interests include federated learning, mobile edge computing, and distributed systems.
\end{IEEEbiography}

\begin{IEEEbiography}[{\includegraphics[width=1in,height=1.25in,clip,keepaspectratio]{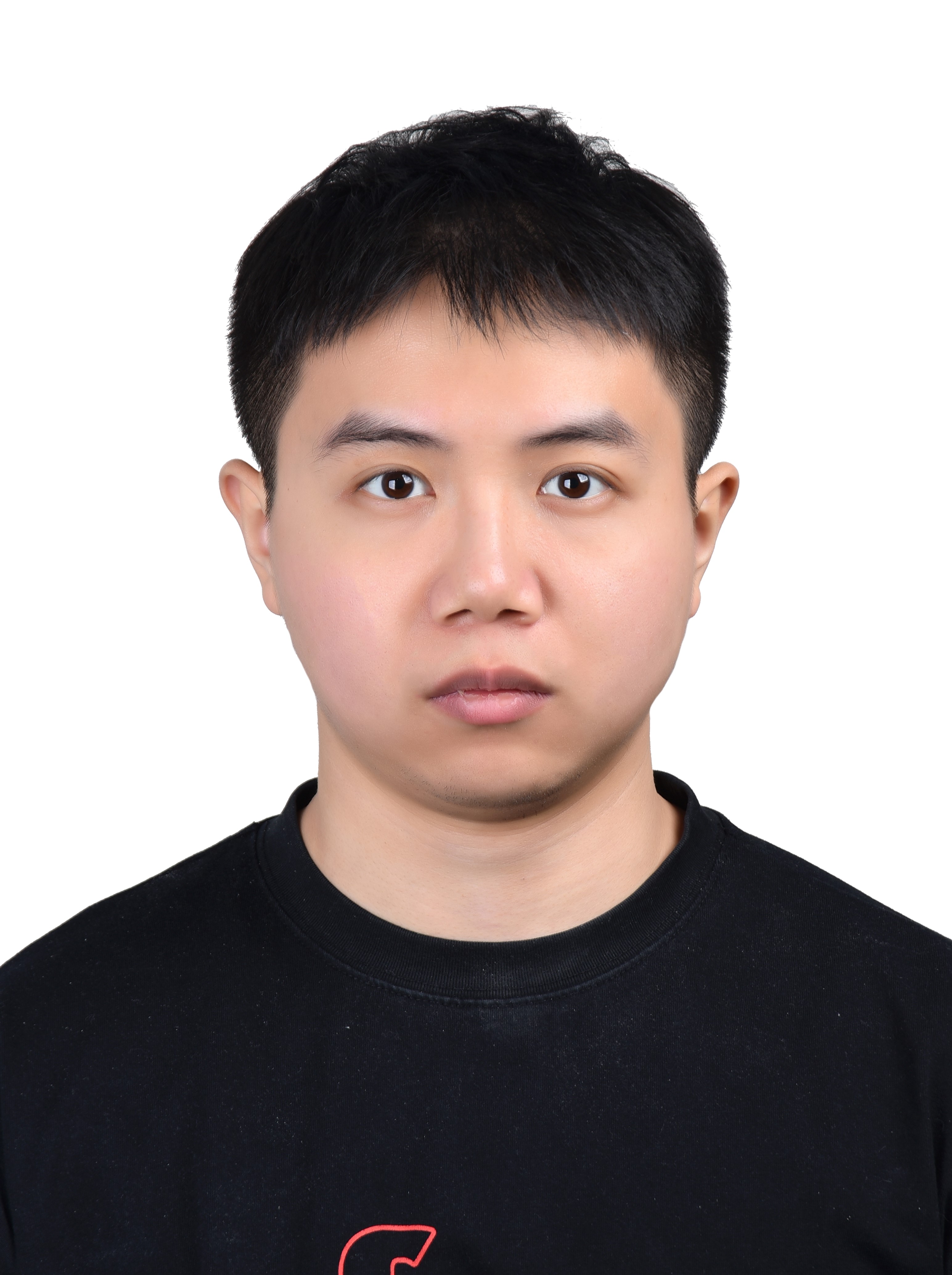}}]{Xujing Li}
(Student Member, IEEE) receives his Ph.D. degree with Institute of Computing Technology, Chinese Academy of Sciences, Beijing, China. His primary research focuses on efficient federated learning and distributed learning optimization. He received his bachelor’s degree at Central South University, Changsha, China in 2019.
\end{IEEEbiography}

\begin{IEEEbiography}[{\includegraphics[width=1in,height=1.25in,clip,keepaspectratio]{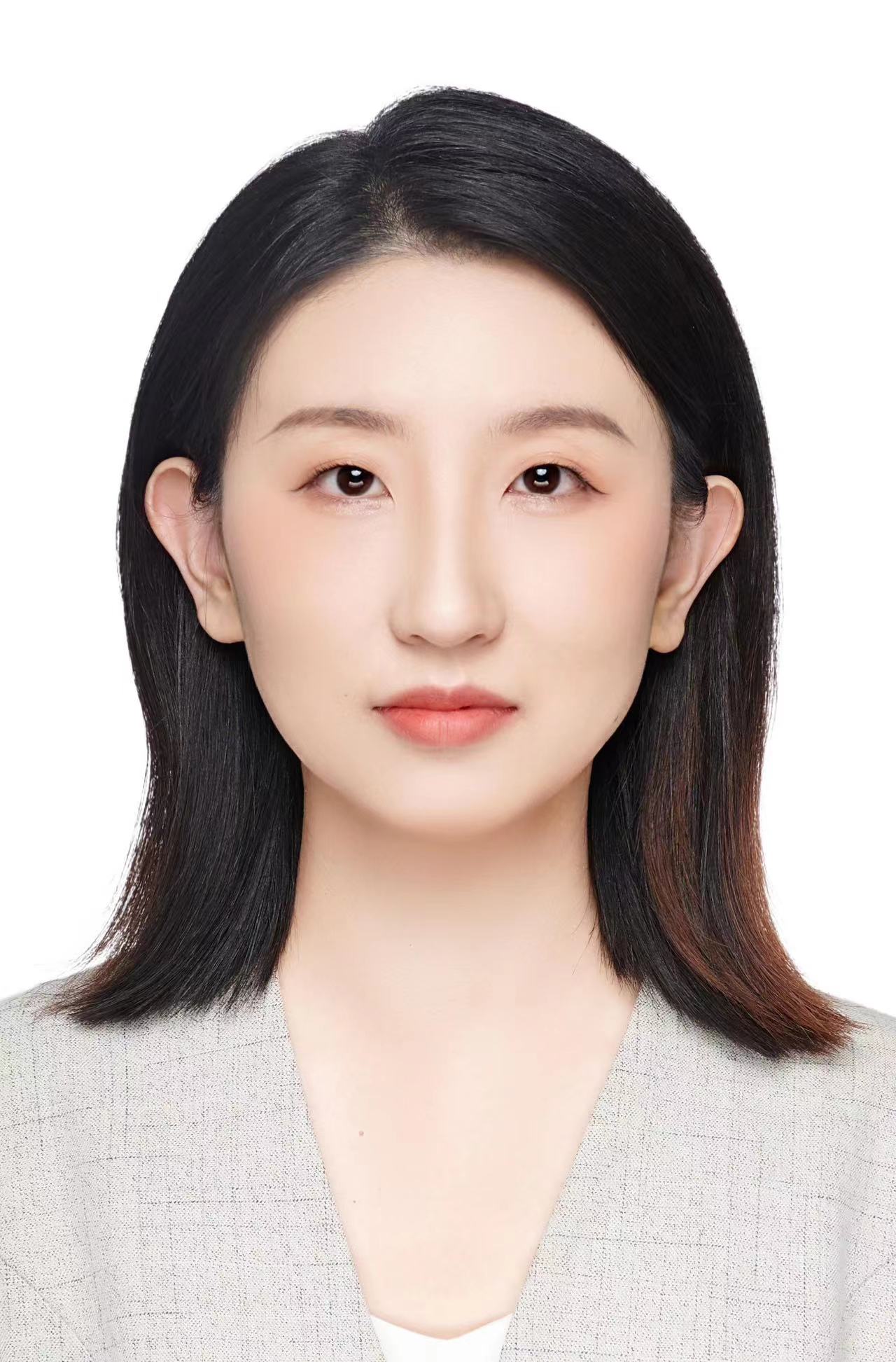}}]{Sheng Sun}
 received her B.S. and Ph.D. degrees
in computer science from Beihang University,
China, and the University of Chinese Academy
of Sciences, China, respectively. She is currently
an associate professor at the Institute of Computing Technology, Chinese Academy of Sciences,
Beijing, China. Her current research interests include federated learning, mobile computing, and edge intelligence.
\end{IEEEbiography}

\begin{IEEEbiography}[{\includegraphics[width=1in,height=1.25in,clip,keepaspectratio]{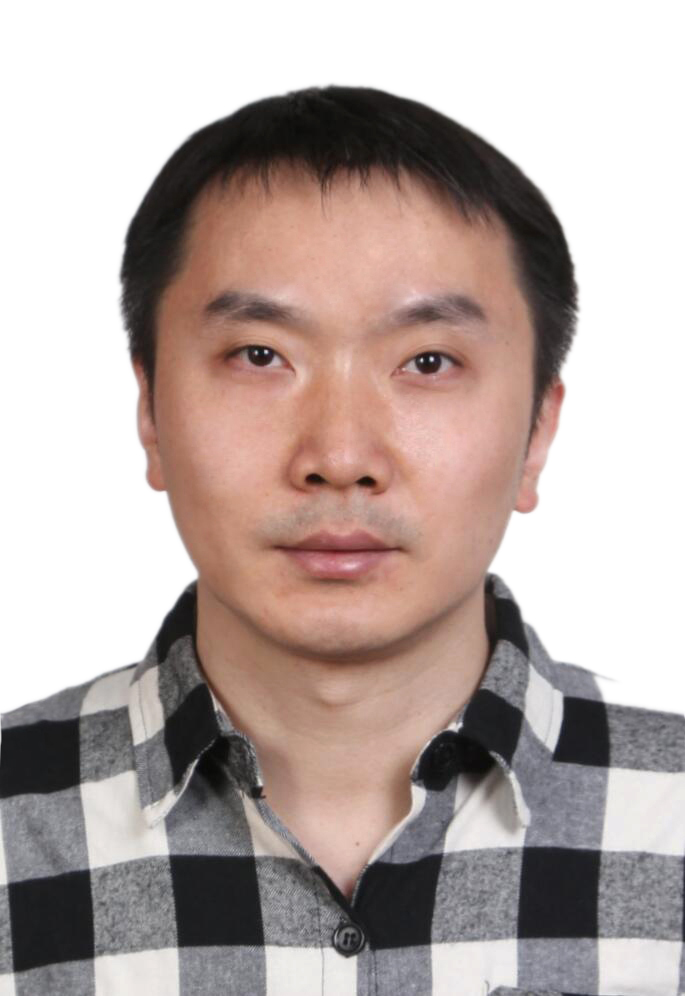}}]{Gang Xu}
received his M.S. degrees in computer science from  the Graduate University of the Chinese Academy of Sciences, China. His current research interests include mobile computing and edge intelligence.
\end{IEEEbiography}

\begin{IEEEbiography}[{\includegraphics[width=1in,height=1.25in,clip,keepaspectratio]{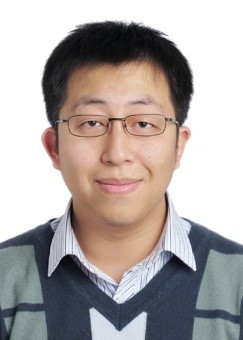}}]{Yuwei Wang}(Member, IEEE) received his Ph.D. degree in computer science from the University of Chinese Academy of Sciences, Beijing, China. He is currently an associate professor at the Institute of Computing Technology, Chinese Academy of Sciences. He has been responsible for setting over 30 international and national standards, and also holds various positions in both international and national industrial standards development organizations (SDOs) as well as local research institutions. His current research interests include federated learning, mobile edge computing, and next-generation network architecture.
\end{IEEEbiography}

\begin{IEEEbiography}[{\includegraphics[width=1in,height=1.25in,clip,keepaspectratio]{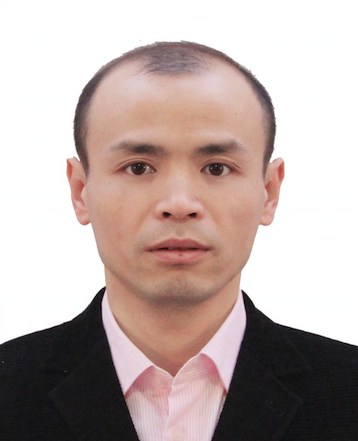}}]{Qi Li} 
(Senior Member, IEEE) received the PhD degree from Tsinghua University. Now, he is an associate professor of Institute for Network Sciences and Cyberspace, Tsinghua University. He has ever worked with ETH Zurich, the University of Texas at San Antonio, the Chinese University of Hong Kong, and the Chinese Academy of Sciences. His research interests include network and system security, particularly in Internet and cloud security, mobile security and big data security. He is currently an editorial board member of the IEEE Transactions on  Dependable and Secure Computing and the ACM Digital Threats: Research and Practice. He is a senior member of the IEEE.
\end{IEEEbiography}

\begin{IEEEbiography}[{\includegraphics[width=1in,height=1.25in,clip,keepaspectratio]{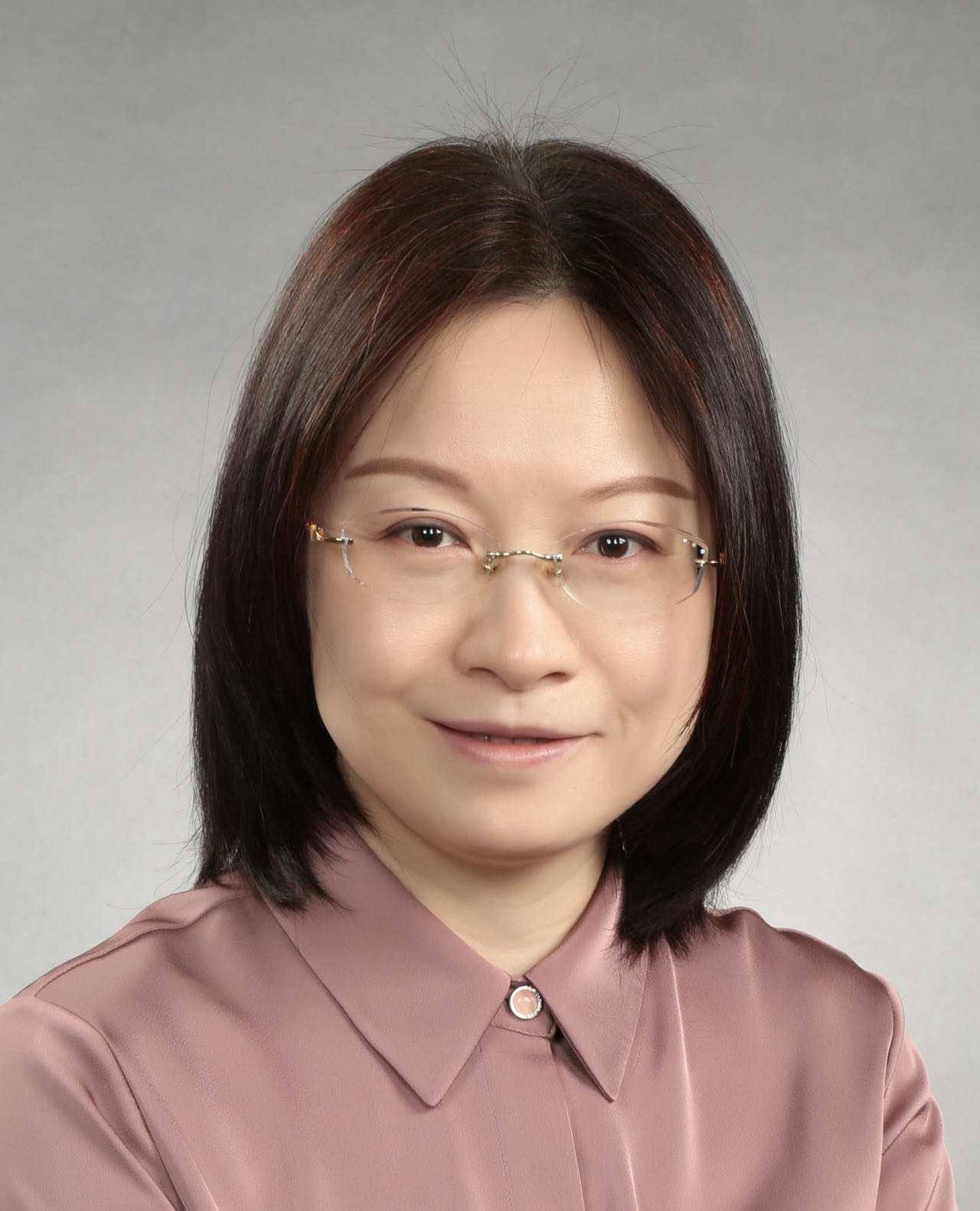}}]{Min Liu} 
(Senior Member, IEEE) received her Ph.D degree in computer science from the Graduate University of the Chinese Academy of Sciences, China. Before that, she received her B.S. and M.S. degrees in computer science from Xi’an Jiaotong University, China. She is currently a professor at the Institute of Computing Technology, Chinese Academy of Sciences, and also holds a position at the Zhongguancun Laboratory. Her current research interests include mobile computing and edge intelligence.
\end{IEEEbiography}

 




\vfill

\end{document}